\documentclass[conference]{IEEEtran}
\usepackage{times}
\pdfoutput=1

\newcommand{\algname}{UWM}

\newcommand{\algnamefull}{Unified World Model}
\newcommand{\algnamefulls}{Unified World Models}
\newcommand{\std}[1]{\tiny{$\pm$ \makebox[8pt][r]{#1}}}
\newcommand{\placeholder}{\makebox[14pt][c]{--}}

\usepackage[numbers]{natbib}
\usepackage{multicol}
\usepackage[bookmarks=true]{hyperref}
\usepackage{booktabs}
\usepackage{graphicx}
\usepackage{xcolor}
\usepackage{bm}
\usepackage{amssymb}
\usepackage{amsmath}
\usepackage{multirow}

\PassOptionsToPackage{numbers, compress}{natbib}

\begin{document}

\title{Unified World Models: Coupling Video and Action Diffusion for Pretraining on Large Robotic Datasets}





%
\author{
\authorblockN{
Chuning Zhu\authorrefmark{2},
Raymond Yu\authorrefmark{2},
Siyuan Feng\authorrefmark{3}, 
Benjamin Burchfiel\authorrefmark{3}, 
Paarth Shah\authorrefmark{3}, and
Abhishek Gupta\authorrefmark{2}}
\authorblockA{\authorrefmark{2}
Paul G. Allen School of Computer Science and Engineering, University of Washington}
\authorblockA{\authorrefmark{3}Toyota Research Institute}
}

\maketitle

\begin{abstract}
Imitation learning has emerged as a promising approach towards building generalist robots. However, scaling imitation learning for large robot foundation models remains challenging due to its reliance on high-quality expert demonstrations. Meanwhile, large amounts of video data depicting a wide range of environments and diverse behaviors are readily available. This data provides a rich source of information about real-world dynamics and agent-environment interactions. Leveraging this data directly for imitation learning, however, has proven difficult due to the lack of action annotation. In this work, we present \algnamefulls{} (\algname{}), a framework that allows for leveraging both video and action data for policy learning. Specifically, a \algname{} integrates an action diffusion process and a video diffusion process within a unified transformer architecture, where \textit{independent} diffusion timesteps govern each modality. By controlling each diffusion timestep, \algname{} can flexibly represent a policy, a forward dynamics, an inverse dynamics, and a video generator. Through simulated and real-world experiments, we show that: (1) \algname{} enables effective pretraining on large-scale multitask robot datasets with both dynamics and action predictions, resulting in more generalizable and robust policies than imitation learning, (2) \algname{} naturally facilitates learning from action-free video data through independent control of modality-specific diffusion timesteps, further improving the performance of finetuned policies. Our results suggest that \algname{} offers a promising step toward harnessing large, heterogeneous datasets for scalable robot learning, and provides a simple unification between the often disparate paradigms of imitation learning and world modeling. Videos and code are available at \url{https://weirdlabuw.github.io/uwm/}

\end{abstract}

\IEEEpeerreviewmaketitle

\begin{figure*}[t!]
    \centering
    \includegraphics[width=0.75\textwidth]{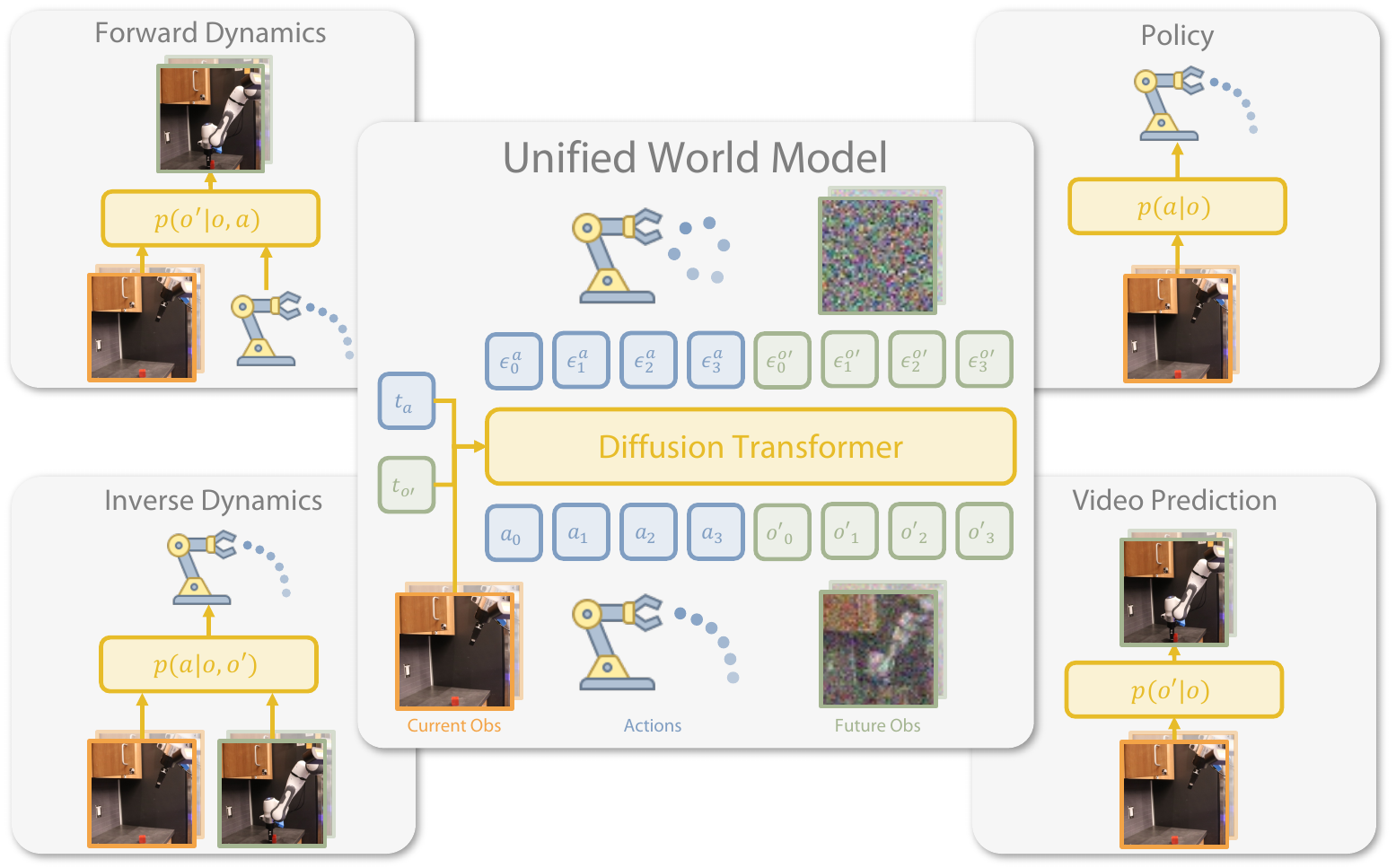} 
    \caption{\algnamefulls{} integrates action and video diffusion in a unified transformer architecture controlled by modality-specific diffusion timesteps. The model can be trained on large robotics datasets and then flexibly perform a variety of different inferences at test time. This enables improved robustness and generalization for imitation learning.}
    \label{fig:teaser}
    \vspace{-0.5cm}
\end{figure*}

\section{Introduction}

Imitation learning provides a simple and scalable way to imbue autonomous robots with complex behaviors using human demonstrations~\cite{pi0, kim24openvla, chi2023diffusionpolicy, act}. Imitation learning via supervised learning, often referred to as behavior cloning (BC), has shown remarkable success due to the advent of powerful multimodal generative models such as diffusion~\cite{ho2020denoising} or flow-based models ~\cite{lipman2023flow}. With these methods, acquiring new behaviors amounts to collecting demonstrations and fitting a generative model to the action distributions given observations. However, despite showing robust and reliable behavior within the training distribution, these methods can be brittle when tasked beyond this distribution. A natural solution to synthesizing robust and generalizable controllers with imitation learning is to scale up the number of high-quality, on-robot demonstrations collected through robotic teleoperation. But this data scaling process is expensive and time-consuming. A natural question arises - \textit{are prevalent methodologies for imitation learning making maximal use of the available large-scale datasets?}

While imitation learning methods learn a mapping from states to optimal actions, they do not explicitly capture temporal dynamics that are naturally present in demonstration trajectories or videos. An alternative paradigm that can leverage such dynamics information is that of world modeling: learning approximate models of how the world changes over time. Commonly instantiated as predicting the future observations given current observations (and actions), world models can be trained from large scale robotic datasets ~\cite{cosmos2025, wu2024ivideogpt}, but also from alternative sources of data such as uncurated ``play" data~\cite{cui2023play} or even action-free data such as videos. A variety of world modeling techniques, such as video diffusion models ~\cite{ho2022video} or latent state-space modeling ~\cite{oshima2024ssm}, have shown impressive results on realistic generation of future frames. However, it is not yet clear how the ability of these world models to capture temporal dynamics can be brought to bear on improving the robustness and generalization of robotic controllers synthesized via imitation learning.  

In this work, we propose a new diffusion-based learning framework that unifies imitation learning and world modeling, incorporating knowledge of temporal dynamics gleaned from large robotic datasets into imitation learning policies. Our key insight is to integrate an action diffusion process and an image diffusion process into a single diffusion transformer model conditioned on \textit{independent diffusion timesteps}. Leveraging a connection between diffusion noise at different timesteps of the forward diffusion noising process and partial masking, this allows for flexible sampling from a number of distributions simply by manipulating the diffusion timesteps independently at inference time. For example, to draw a sample from the policy, one can ``mask out" the image diffusion process by fixing the image diffusion timestep to $T$. Similarly, one can sample from the forward dynamics model by fixing the action diffusion timestep to $0$, inferring next observations given current observations and  ``clean" actions. Same holds for inverse dynamics models and unconditional video prediction models into the future. This yields a simple, unified diffusion model that can serve as a policy, dynamics model, video predictor or inverse model (Fig~\ref{fig:teaser}). 

Concretely, a \algname{} consists of a coupled score model that predicts action scores and \emph{future} image scores, conditioned on the current image and separate diffusion timesteps for action and future image. During training, the timesteps are sampled independently at random, exposing the model to different combinations of action and image noises. During inference, \algname{} enables flexible sampling from various distributions by manipulating the diffusion timesteps independently. In particular, a \algname{} can generate samples from (1) forward dynamics, (2) inverse dynamics (3) marginal action distribution (policy), (4) marginal image distribution (video generative model). We show that this learning framework leads to improved policies compared to standard imitation learning since, (1) the unified architecture enables feature sharing between actions and pixels, resulting in additional supervision from the same data, (2) the model captures all combinations of marginal and conditional distributions, acquiring an understanding of the causal relationship between actions and images, (3) the model can learn from broader data modalities such as action-free videos.

We demonstrate the effectiveness of \algname{} through a set of experiments across both simulation and real-world robotic manipulation tasks. We show that \algname{} is capable of extracting knowledge from multitask robotic datasets, and further leveraging action-free video data to improve its generalization to out-of-distribution conditions. These models are able to flexibly perform a variety of test-time inference, while retaining strong performance of both policy and dynamics prediction. Through this investigation of \algname{}, we take a step towards bridging the gap between policies and world models for robot learning.
\section{Preliminaries}
\label{sec:prelims} 
Unified world models build on the framework of denoising diffusion probabilistic models ~\cite{ho2020denoising} and their application to problems in robotic control ~\cite{chi2023diffusionpolicy}. 

\subsection{Diffusion Models}

Denoising Diffusion Probabilistic Models (DDPMs)~\cite{ho2020denoising} are a family of generative models that define a forward noising process and a learned reverse denoising process to generate samples from a complex, multimodal data distribution. Let $p(x_0)$ denote the data distribution from which a number of samples are available. In the \textit{forward} diffusion process, the  data $x_0 \sim p(x_0)$ is gradually corrupted by iteratively adding Gaussian noise over $T$ steps through a Markov chain according to a variance schedule $\{\beta_t\}_{t=1}^T$. Concretely, the forward process is defined as
$$
q(x_{t} \mid x_{t-1}) = \mathcal{N}\left(x_t \mid \sqrt{1 - \beta_t}\,x_{t-1},\,\beta_t\,I\right),\quad t=1,\dots,T.
$$
After $T$ steps, $x_T$ is nearly an isotropic Gaussian~\cite{ho2020denoising}. The corresponding \textit{reverse} process aims to map $x_T$ back to a clean sample $x_0$ from the data distribution by iteratively denoising. While the exact reverse conditional $q(x_{t-1} \mid x_t)$ is generally intractable, one can learn a parametrized approximation,
$$
p(x_{t-1} \mid x_t)
=
\mathcal{N}\bigl(\mu(x_t,\,t),\,\Sigma(x_t,\,t)\bigr),
$$
In practical settings, the variance $\Sigma(x_t,\,t)$ is set to a simple time varying constant $\sigma_t^2 I$. As shown in prior work~\cite{bao2022analytic}, the optimal mean under MLE is: 
$$
\mu(x_t, t) = \frac{1}{\sqrt{\alpha_t}}\Bigg(x_t - \frac{\beta_t}{\sqrt{1 - \bar{\alpha}_t}}\mathbb{E}\left[ \epsilon_x | x_t \right]\Bigg)
$$
where $\alpha_t = 1 - \beta_t$ and $\bar{\alpha}_t = \prod_{i=1}^t \alpha_i$ and $\epsilon_x$ is the Gaussian noise injected into $x_t$. To approximate this conditional expectation $\mathbb{E}\left[ \epsilon_x | x_t \right]$, DDPMs train a neural network $s_\theta$ using a variant of \textit{denoising score matching}:
$$
\min_\theta 
\;
\mathbb{E}_{x_0,\,t,\,\epsilon}\Bigl[\|\,s_\theta(x_t,\,t) - \epsilon\,\|_2^2\Bigr],
$$
where $x_t = \sqrt{\bar{\alpha}_t}\,x_0 + \sqrt{1-\bar{\alpha}_t}\,\epsilon$ and $\bar{\alpha}_t = \prod_{i=1}^t (1 - \beta_i)$. Intuitively, this ``score function" predicts the noise added at each step using a simple regression objective. This learned noise prediction network $s_\theta(x_t,\,t)$ can then directly parameterize the reverse diffusion process as
$$
p_\theta(x_{t-1} \mid x_t)
=
\mathcal{N}\bigl(
\frac{1}{\sqrt{\alpha_t}}(x_t - \frac{\beta_t}{\sqrt{1 - \bar{\alpha}_t}}s_\theta(x_t,\,t)), \sigma_t^2 I \bigr)
$$

Given this reverse diffusion model, samples can approximately be drawn from the data distribution using a simple ``denoising procedure". Starting with a sample $x_T \sim \mathcal{N}(0,\,I)$ drawn from Gaussian noise, new samples are iteratively drawn from $p_\theta(x_{t-1}\mid x_t)$ until a ``clean sample" $x_0$ is obtained. This procedure allows for the representation of complex multimodal distributions where performing MLE tractably is challenging.

\paragraph{Conditional Generation with Diffusion Models} While the above-mentioned generative modeling process is unconditional, diffusion models can be naturally extended to conditional settings. Consider a setting where multivariate data $(x_0, z_0) \sim p(x_0, z_0)$ is available, and the conditional distribution $p(x_0|z_0)$ must be modeled. In these settings, a bulk of the machinery from above can be reused, simply with an additional conditioning variable. The forward process remains identical, while the reverse process is modified as 

$$
p(x_{t-1} \mid x_t, z_0)
=
\mathcal{N}\bigl(
\frac{1}{\sqrt{\alpha_t}}(x_t - \frac{\beta_t}{\sqrt{1 - \bar{\alpha}_t}}\mathbb{E}\left[ \epsilon_x | x_t, z_0 \right]), \sigma_t^2 I \bigr)
$$

In this case, the expectation $\mathbb{E}\left[ \epsilon_x | x_t, z_0 \right]$ can be approximated using a \emph{conditional} noise prediction network that is also trained with denoising score matching:
$$
\min_\theta 
\;
\mathbb{E}_{(x_0,z_0) \sim p(x_0, z_0)\,t,\,\epsilon}\Bigl[\|\,s_\theta(x_t, z_0, \,t) - \epsilon\,\|_2^2\Bigr],
$$
\section{Method}

In this section, we introduce Unified World Models as a way to incorporate temporal dynamics into diffusion-based action prediction models, proving a bridge between the often disparate worlds of imitation learning and world modeling. 

\subsection{Problem Setup}
We build on typical sequential decision making settings, assuming access to a dataset of (observation, action, next-observation) pairs $\mathcal{D}_{\text{e}} = \{ (o_i, a_i, o'_{i}) \}_{i=1}^N $ provided by an expert demonstrator. For the sake of exposition, we will assume that the environment is Markovian in observations $o$. In addition to this action-labeled dataset, we may also have access to an action-free dataset  $\mathcal{D}_{\text{af}} = \{ (o_i,o_{i+1}) \}_{i=1}^M $. The question becomes - how can we extract the most learning signal out of these datasets for synthesizing robot controllers?

In this context, several different models may be desired: (1) a policy $p(a|o)$ (often referred to as $\pi(a|o)$) that samples optimal actions to execute at a particular observation, (2) a dynamics model $p(o'|o, a)$ that samples future observations, given a current observation and action, (3) an inverse model $p(a|o, o')$ that predicts what distribution of actions can transition between a current observation and a desired next observation, and (4) a video prediction model $p(o'|o)$ that predicts marginal future observations given current ones. While these models have each seen use in different contexts, they are largely considered to be disparate fields of study. In this work, we show that these are many sides of the same dice; they can be unified into a single model to benefit each other.

\subsection{Unified World Models via Coupled Video-Action Diffusion}
\label{sec:uwm}
\begin{figure*}[t!]
    \centering
    \includegraphics[width=\textwidth]{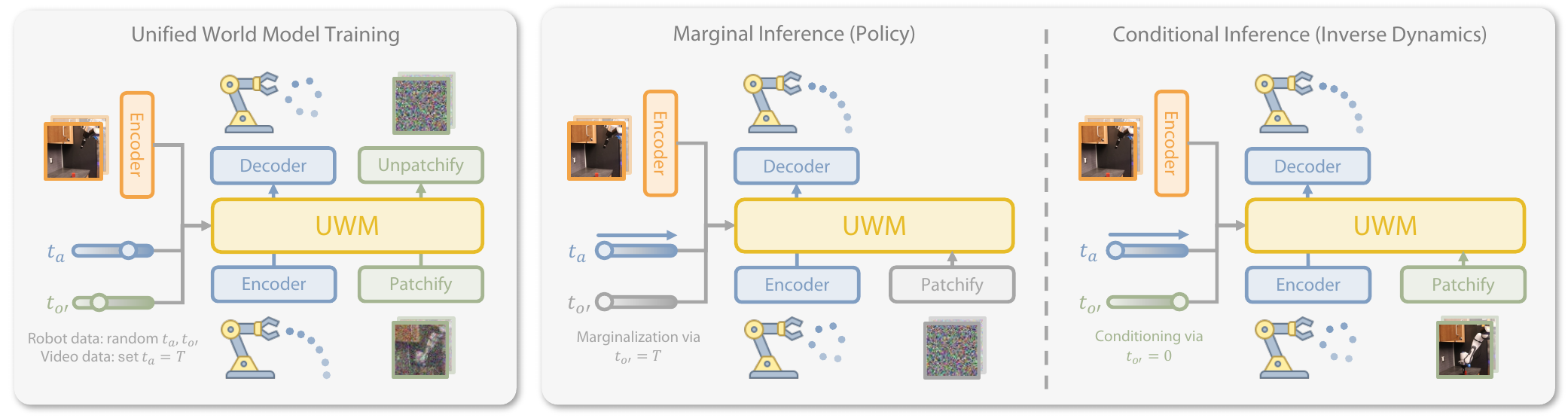} 
    \caption{Unified World Model training and inference pipeline. The left panel shows \algname{} pretraining on robot trajectories with actions and co-training on action-free videos by masking out actions using diffusion timesteps. The right panel illustrates marginal and conditional inference modes, corresponding to the policy and the inverse dynamics.}
    \label{fig:method}
\end{figure*}

The core idea of a \algname{} is to develop a single diffusion model that can be trained on samples from the joint distribution of data $p(o, a, o')$ and used to flexibly perform inference for the policy $p(a|o)$, the dynamics model $p(o'|o, a)$, the inverse model $p(a|o, o')$ and the video prediction model $p(o'|o)$, with minimal modifications to test-time inference.

We start by instantiating a joint diffusion model that integrates next observation prediction $o'$ and action prediction $a$ into a single diffusion model conditioned on current observation $o$. This can naturally be done by parameterizing a joint noise prediction network $s_\theta(o, a_{t}, o'_{t}, \,t)$ that approximates a conditional expectation over both action and next observation noise $\mathbb{E}\left[  \epsilon_{a}, \epsilon_{o'} | o, a_t, o'_t, \right]$, with $\epsilon_{a}, \epsilon_{o'}$ referring to noise on actions and next observations, and $t$ referring to the \emph{coupled} timestep of the joint diffusion process.\footnote{It is important to note that throughout this section $o$ will refer to the current observation, $o'_t$ will refer to timestep $t$ in the diffusion process for the \textbf{next} observation, and $a_t$ will refer to timestep $t$ in the diffusion process for action.} However, training such a joint noise prediction network~\cite{pad} $s_\theta(o, a_{t}, o'_{t}, \,t)$ does not accomplish flexible inference since it can only sample from the joint distribution of $(a, o')$.

For flexible inference, we can leverage a connection between diffusion time-steps and masking - \emph{noising input tokens by setting the inference timestep for diffusion appropriately can naturally induce a form of partial masking}. Time-steps closer to $T$ (fully noised) indicate full masking, while timesteps closer to $0$ (unnoised) indicate no masking. Based on this key insight, \algname{} modifies the joint diffusion process mentioned above and \emph{decouples} the timesteps between that of the diffusion processes of next observation prediction $t_{o'}$ and that of action prediction $t_a$ in a joint noise prediction network $s_\theta$. This separation of time steps allows for independent control of $t_{o'}$ and $t_a$ during training and inference, which gives rise to flexible inference capabilities.

\newcommand{\noiseprednet}{$s_\theta(o, a_{t_a}, o'_{t_{o'}}, t_a, t_{o'})$}
Concretely, a \algname{} models a coupled noise prediction network \noiseprednet{} that approximates a conditional expectation over noise $\mathbb{E}\left[ \epsilon_{a}, \epsilon_{o'} | o, a_{t_a}, o'_{t_{o'}}\right]$, with $\epsilon_{a}, \epsilon_{o'}$ referring to noise on next observations and actions, and $t_a$ and $t_{o'}$ referring to the \emph{decoupled} steps of the diffusion process with respect to actions and next-observations respectively. The ability to set diffusion timesteps independently allows for marginalization and conditioning of different variables. Fixing the timestep for either $t_a$ or $t_{o'}$ to $T$ marginalizes the corresponding variable $a$, $o'$, while setting the timestep to $0$ performs conditioning. By setting timestep $t_{o'} = T$, the joint model is approximating the expectation $\mathbb{E}\left[ \epsilon_{a}, \epsilon_{o'}, | o, a_{t_a}, o'_{T} \right]$. Since $o'_{T}$ is approximately an isotropic Gaussian, this reduces to $\mathbb{E}\left[ \epsilon_{a} |o, a_{t_a} \right]$, which represents a policy $p(a|o)$, thereby performing marginalization. Similarly, setting the timestep $t_{o'} = 0$ reduces the approximated distribution to $\mathbb{E}\left[ \epsilon_{a} |o, a_{t_a}, o' \right]$ which corresponds to an inverse model $p(a|o, o')$, thereby performing conditioning. Simply setting combinations of $t_a$ and $t_{o'}$ allows for flexible inference of policies, dynamics models, inverse models, and video prediction from the same model.

This suggests a training recipe using a simple modification to the standard denoising objective~\cite{ho2020denoising}. To train a joint noise prediction diffusion model $(\epsilon^\theta_a, \epsilon^\theta_{o'}) = s_\theta(o'_{t_{o'}}, a_{t_a}, o, t_a, t_{o'})$, we independently sample action timestep $t_a$ and next observation timestep $t_{o'}$, draw noisy action and next-observation samples from their respective distributions, and train the coupled conditional score model conditioned on the current observation with a standard denoising objective across both actions and next-observations: 
\begin{align}
\ell(\theta) = \mathbb E_{\substack{(o,a,o') \sim \mathcal{D} \\ t_a,t_{o'} \sim \mathcal{U}(0, T) \\ \epsilon_a,\epsilon_{o'} \sim \mathcal{N}(0,1)}}  &\Bigl[w_a \parallel \epsilon^\theta_a - \epsilon_a \parallel_2^2 + w_{o'} \parallel \epsilon^\theta_{o'} - \epsilon_{o'} \parallel_2^2\Bigr], \label{eq:objective}\\
\textit{where} \quad \epsilon^\theta_a, \epsilon^\theta_{o'} &= s_\theta(o, a_{t_a}, o'_{t_{o'}}, t_a, t_{o'}), \nonumber\\
a_{t_a} &= \sqrt{\bar \alpha_{t_a}}a + \sqrt{1 - \bar\alpha_{t_a}} \epsilon_a, \nonumber\\
o'_{t_{o'}} &= \sqrt{\bar \alpha_{t_{o'}}}o' + \sqrt{1 - \bar\alpha_{t_{o'}}}\epsilon_{o'}. \nonumber
\end{align}
where $w_a$ and $w_{o'}$ are weights chosen to trade off between the action prediction and next-observation prediction objectives. Intuitively, this training paradigm exposes the model to all combinations of noise levels of the modalities. At inference, we can flexibly draw samples from various distributions by controlling the timesteps $t_a$ and $t_{o'}$ as follows:

\begin{enumerate}
\item \textbf{Policy} To sample from the policy $p(a|o)$, we \emph{marginalize} out the next observation $o'$ by setting $t_{o'} = T$ and $o'_T \sim \mathcal{N}(0, I)$. We perform the reverse diffusion process on actions going from $t_a = T, \dots, 1$ with $a_T \sim \mathcal{N}(0, I)$: 

\begin{equation}
\begin{aligned}
    a_{t-1} &= \frac{1}{\sqrt{\alpha_t}} 
    \bigg(a_t - \frac{\beta_t}{\sqrt{1 - \bar{\alpha}_t}} 
    s_\theta(o, a_{t}, o'_{T}, t, T) \bigg) \\
    &\quad + \sigma_t \delta_t, \quad\quad \delta_t \sim \mathcal{N}(0, I)
\end{aligned}
\end{equation}

\item \textbf{Video Prediction Model} To sample from the video prediction model $p(o'|o)$, we \emph{marginalize} out the action $a$ by setting $t_{a} = T$ and $a_T \sim \mathcal{N}(0, I)$. We perform the reverse diffusion process on next observations going from $t_{o'} = T, \dots, 1$ with $o'_T \sim \mathcal{N}(0, I)$:

\begin{equation}
\small
\begin{aligned}
    o'_{t-1} &= \frac{1}{\sqrt{\alpha_t}} 
    \bigg( o'_t - \frac{\beta_t}{\sqrt{1 - \bar{\alpha}_t}} 
    s_\theta(o, a_{T}, o'_{t}, T, t) \bigg) \\
    &\quad + \sigma_t \delta_t, \quad\quad \delta_t \sim \mathcal{N}(0, I)
\end{aligned}
\end{equation}

\item \textbf{Forward Dynamics} To sample from the forward dynamics model $p(o'|o, a)$, we \emph{condition} on a particular action $a$ by setting $t_{a} = 0$ and $a_0 = a$. We perform the reverse diffusion process on next observations going from $t_{o'} = T, \dots, 1$ with $o'_T \sim \mathcal{N}(0, I)$: 

\begin{equation}
\begin{aligned}
    o'_{t-1} &= \frac{1}{\sqrt{\alpha_t}} 
    \bigg( o'_t - \frac{\beta_t}{\sqrt{1 - \bar{\alpha}_t}} 
    s_\theta(o, a, o'_{t}, 0, t) \bigg) \\
    &\quad + \sigma_t \delta_t, \quad\quad \delta_t \sim \mathcal{N}(0, I)
\end{aligned}
\end{equation}

\item \textbf{Inverse Dynamics}
To sample from the inverse dynamics model $p(a|o, o')$, we \emph{condition} on a particular next observation $o'$ by setting $t_{o'} = 0$ and $o_0 = o$. We perform the reverse diffusion process on actions going from $t_{a} = T, \dots, 1$ with $a_T \sim \mathcal{N}(0, I)$: 

\begin{equation}
\begin{aligned}
    a_{t-1} &= \frac{1}{\sqrt{\alpha_t}} 
    \bigg( a_t - \frac{\beta_t}{\sqrt{1 - \bar{\alpha}_t}} 
    s_\theta(o, a_{t}, o', t, 0) \bigg) \\
    &\quad + \sigma_t \delta_t, \quad\quad \delta_t \sim \mathcal{N}(0, I)
\end{aligned}
\end{equation}
\end{enumerate}

This simple modification to the standard diffusion training paradigm allows a single model to be trained, benefiting from feature sharing between different models of action and future observation prediction. This model can then be flexibly used for inference with just the choice of timesteps, making it a versatile, general-purpose decision-making model.

\subsection{Architecture}
\label{sec:practical}
\begin{figure}[t]
    \centering
    \includegraphics[width=0.8\linewidth]{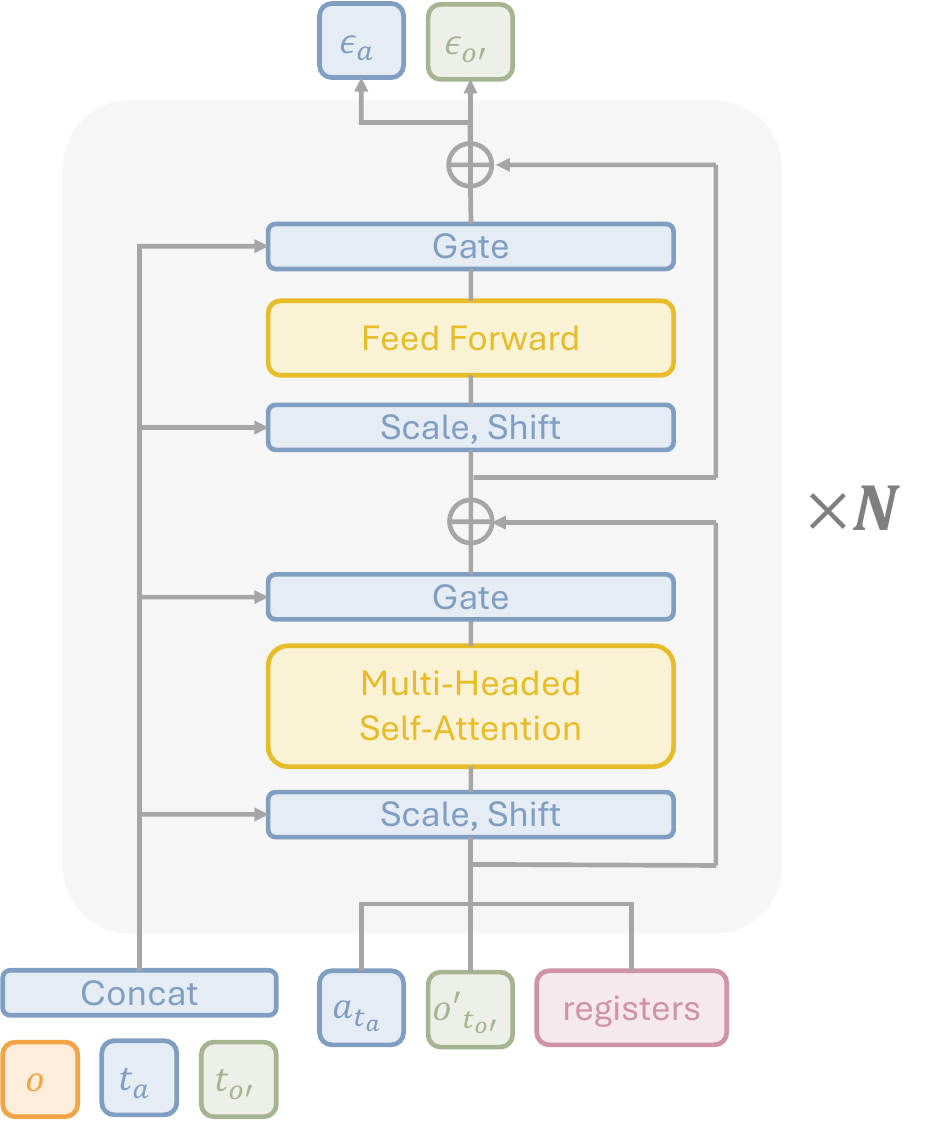} 
    \caption{A single \algnamefull{} (\algname{}) block consists of a transformer block with observations and diffusion timesteps conditioning via adaptive layer norm. In addition, we add randomly initialized register tokens which allows for better multi-modal feature sharing.}
    \label{fig:architecture}
\end{figure}

We model the \algname{} as a diffusion transformer as shown in Fig. \ref{fig:method} and \ref{fig:architecture}. The model predicts actions and observation noises $\epsilon_{a}$ and $ \epsilon_{o'}$ given current observations $o$, noisy actions $a_{t_a}$, noisy observations $o'_{t_{o'}}$, action timestep $t_a$, and observation timestep $t_o$. The actions are action chunks of length $h_a$. The current observations $o$ and next observations $o'$ are frame-stacked observations of length $h_o$ from $n_c$ camera views. 

To condition the model on current observations, we encode each frame from each camera view using a ResNet-18~\cite{He2015DeepRL} encoder to obtain an $n_{\text{embd}}$ dimension feature. the features are concatenated to form an embedding of size $n_c \cdot h_o \cdot n_{\text{embd}}$. The diffusion timesteps are encoded using a shared sinusoidal timestep encoder from \cite{ho2020denoising}, resulting in two timestep embeddings. These timestep embeddings are concatenated with the image features, and the combined features are used to condition the transformer via Adaptive Layer Normalization (AdaLN) ~\cite{Peebles2022DiT}. 

The context of the diffusion transformer consists of action embeddings and image embeddings. The action embeddings are obtained by encoding the action chunk per-timestep using a shallow MLP. For image diffusion, we adopt the latent diffusion paradigm \cite{rombach2021highresolution} and encode full-size $(224, 224, 3)$ images into $(28, 28, 4)$ latent images using a frozen SDXL VAE \cite{podell2024sdxl}. We then patchify the latent images using a spatiotemporal patchifier of size $(4, 4, 2)$. These image patch embeddings are then concatenated with the action embeddings and passed into the transformer backbone. The image noising and denoising processes are performed in the latent space, and the final image sample is decoded using the same VAE to generate full-size images.

Empirically, we found that adding redundant tokens that are eventually discarded (i.e. registers \cite{darcet2024vision}) helps with model performance. We hypothesize that this is because images and actions are distinct modalities that can benefit from having an intermediary medium to exchange information. However, since all output embeddings of the diffusion transformer are meaningful noise predictions, there is no room for such communication. The registers can store information from either modality, which can then be retrieved in subsequent transformer layers. We demonstrate the effectiveness of registers in our ablation experiments in Section \ref{sec:exp-ablation}.

\subsection{Training Paradigms}
In this work, we evaluate the effectiveness of \algname{} as a pretraining method for learning the dynamics information from large multitask robotic datasets. To train a \algname{} on robot data, we sample sequences of observations and actions from the dataset, construct $(o, a, o')$ tuples, sample random diffusion timesteps $t_a, t_{o'} \sim \mathcal U(0, T)$, and optimize the denoising score matching objective in Eq. \ref{eq:objective}.

Moreover, \algname{} naturally enables co-training on action-free video data by using diffusion timesteps for masking. Given action-free video samples, instead of sampling the action timestep randomly, we fix the action timestep to $T$, impute the missing actions with random noise $\epsilon_a \in \mathcal N(0, 1)$, optimize the same loss in Eq. \ref{eq:objective}. We validate the effectiveness of co-training on videos in our experiments in Section. \ref{sec:exp-real}.
\section{Experiments}

\begin{figure}[t]
    \centering
    \includegraphics[width=\linewidth]{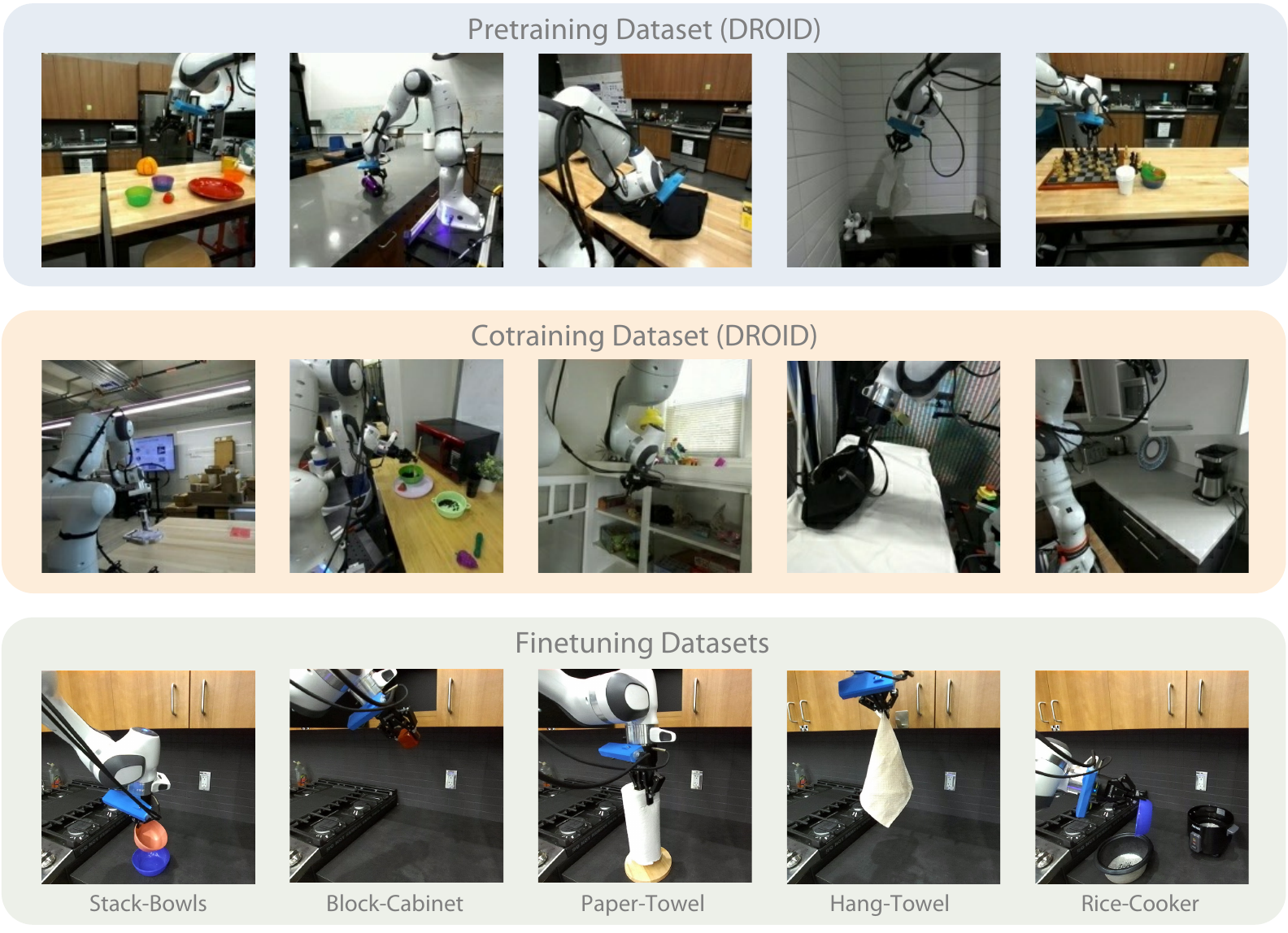} 
    \caption{Visualization of datasets used for pretraining and finetuning. The pretraining and cotraining dataset consists of diverse tasks performed by Franka robots in various environments to ensure broad generalization capabilities. The finetuning datasets include five tasks, each designed to evaluate task-specific performance under controlled conditions.}
    \label{fig:env-droid}
\end{figure}

\begin{figure*}[t]
    \centering
    \includegraphics[width=1.0\linewidth]{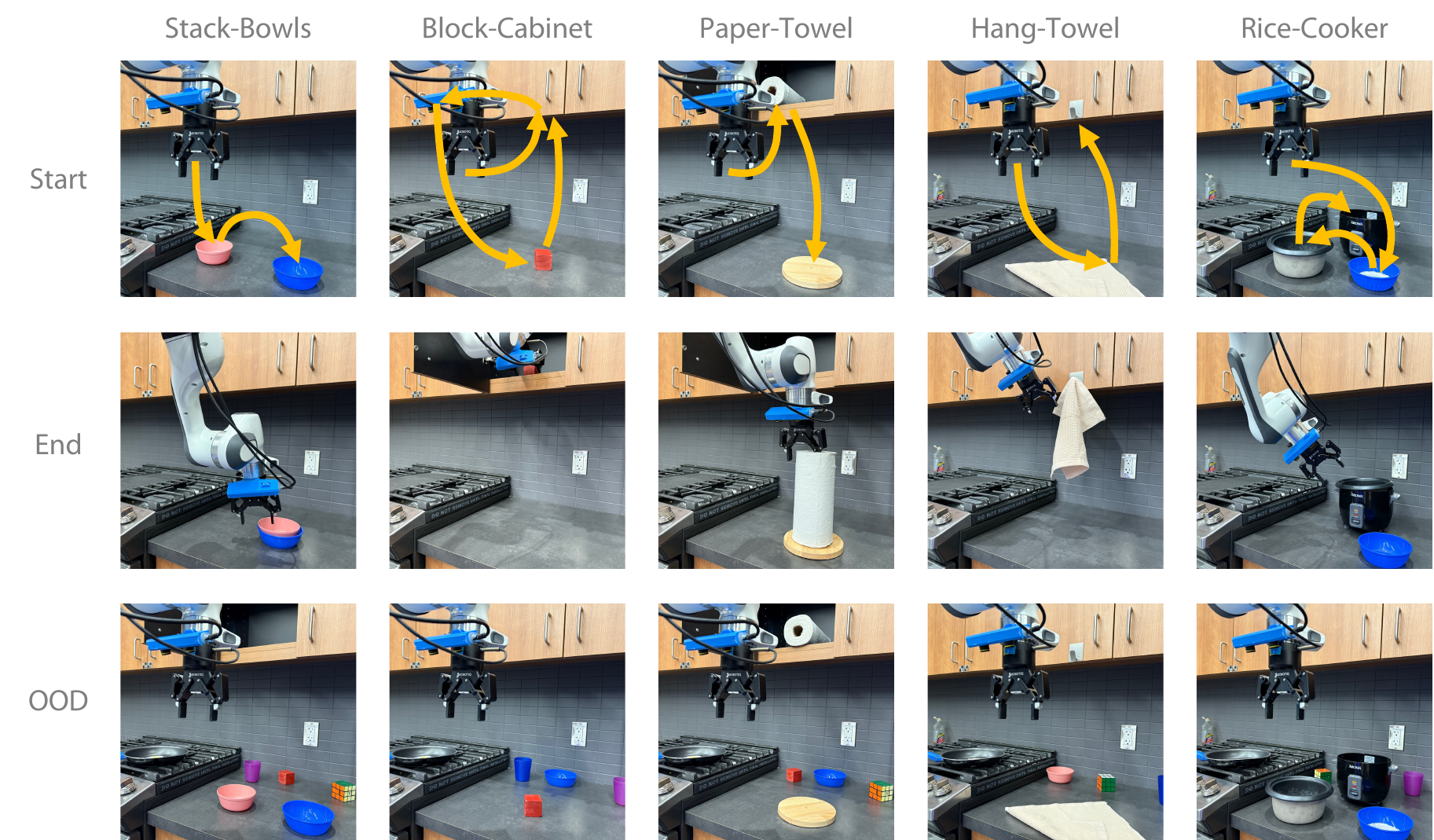} 
    \caption{Setup for real robot tasks: Stack-Bowls, Block-Cabinet, Paper-Towel, Hang-Towel, and Rice-Cooker. The first row illustrates the initial configurations for each task. The second row demonstrates successful task completions. The third row highlights the out-of-distribution (OOD) configurations designed to evaluate the robustness of each method.}
    \label{fig:eval_setup}
    \vspace{-0.2cm}
\end{figure*}

In this section, we examine the following research questions: (1) can \algname{} effectively learn from large robotic datasets as a pretraining paradigm? (2) can \algname{} further benefit from additional video data without action labels in a co-training paradigm? (3) what are the key design choices that contribute to \algname{}'s performance. We answer these questions through a number of real robot experiments with a Franka robot using the DROID ~\cite{droid} manipulation platform, as well as simulated experiments in the LIBERO \cite{liu2023libero} benchmark.

\subsection{Baselines}
We compare \algname{} to the following baselines throughout our experiments. Detailed descriptions of each baseline are deferred to Appendix \ref{app:implementation}.
\begin{enumerate}
\item \textbf{Diffusion Policy (DP)}~\cite{chi2023diffusionpolicy} is a behavior cloning method that fits a conditional diffusion model to a dataset of expert observation-action data. We extend the framework to the pretraining-finetuning setting by fitting a model to the behavior distribution of a multitask dataset and then finetuning it to the task-specific demonstrations. We compare to DP as a baseline to validate the effectiveness of the additional supervisory signals in \algname{}. To minimize the discrepancy from \algname{}, we adopt a diffusion transformer backbone similar to \cite{ditpolicy} instead of the original UNet architecture \cite{chen2024diffusionforcingnexttokenprediction}.
\item \textbf{PAD}~\cite{guo2024prediction} is a video-action diffusion model that learns a joint distribution of actions and future observations conditioned on current observations. The key conceptual difference between PAD and \algname{} is the decoupling of timesteps between actions and next-observations. In addition, PAD conditions the model on the current observations by concatenating the clean latents of the current observations to the noisy latents of the next observations along the channel dimension, similar to~\cite{blattmann2023stablevideodiffusionscaling}. PAD supports co-training on videos by masking the action tokens with a learned mask token. 
\item \textbf{GR1}~\cite{wu2024unleashing} is a video-action transformer model that predicts actions and future image observations conditioned on current image observations. Unlike other baselines, GR1 does not model a distribution over data using a diffusion process. Instead, it directly regresses the actions and images by minimizing a least squares loss. We compare to GR1 to validate the effectiveness of diffusion as a pretraining objective relative to regression. GR1 supports co-training on videos by masking the action tokens with a learned mask token. 
\end{enumerate}

\begin{table*}[t]
\centering
\caption{Evaluation Results Across Real Robot Tasks (Pretrain / Cotrain)}
\begin{tabular}{lccccccccc}
\toprule
& \multicolumn{2}{c}{Stack-Bowls} & \multicolumn{2}{c}{Block-Cabinet} & \multicolumn{2}{c}{Paper-Towel} & \multicolumn{2}{c}{Hang-Towel} & \multicolumn{1}{c}{Rice-Cooker} \\
(Pretrain / Cotrain) & ID & OOD & ID & OOD & ID & OOD & ID & OOD & ID \\
\midrule
UWM (Ours)         & \textbf{0.86 / 0.92} & \textbf{0.76 / 0.84} & \textbf{0.76 / 0.84} & \textbf{0.60 / 0.72} & \textbf{0.78 / 0.86} & \textbf{0.78 / 0.84} & \textbf{0.82 / 0.86} & \textbf{0.64 / 0.76} & \textbf{0.60 / 0.65} \\
DP                 & 0.48 / \placeholder{} & 0.36 / \placeholder{} & 0.60 / \placeholder{} & 0.26 / \placeholder{} & 0.52 / \placeholder{} & 0.48 / \placeholder{} & 0.64 / \placeholder{} & 0.28 / \placeholder{} & 0.35 / \placeholder{} \\
PAD                &  0.08 / 0.20 & 0.08 / 0.12 & 0.00 / 0.00 & 0.00 / 0.00 & 0.42 / 0.42 & 0.34 / 0.44  & 0.52 / 0.54 & 0.30 / 0.38 & 0.00 / 0.00 \\
GR1                & 0.66 / 0.62 & 0.48 / 0.38 &  0.66 / 0.74 & 0.44 / 0.64 & 0.60 / 0.46 & 0.60 / 0.46 & 0.66 / 0.66 & 0.48 / 0.44 &  0.40 / 0.25 \\
\bottomrule
\end{tabular}
\label{tab:combined_results}
\end{table*}

\subsection{Real Robot Experiments}
\label{sec:exp-real}

\subsubsection{Setup}
To evaluate \algname{} and baselines as pretraining methods, we leverage the DROID dataset~\cite{droid} as a source of pretraining data. The DROID dataset is a diverse dataset consisting of robot trajectories collected across various institutions and operators, covering a large variety of tasks, camera positions and backgrounds in natural settings. We curate a pretraining dataset by sampling a subset of 2000 trajectories from the DROID dataset based on location (Fig \ref{fig:env-droid}, top row). For methods that support co-training on video data (e.g. GR-1, PAD, and UWM), we additionally evaluate their capability of learning from action-free videos. To this end, we sample another 2000 trajectories from the rest of the DROID dataset and remove their action annotations to use as videos (Fig \ref{fig:env-droid}, middle row). 

To evaluate the efficacy of the pretrained models, we construct five different real-world tasks (shown in Fig~\ref{fig:eval_setup}) using the portable manipulation platform proposed in DROID~\cite{droid}. The tasks involve different kinds of robotic manipulation:
\begin{itemize}
    \item \textbf{Stack-Bowls} involves picking up the pink bowl and stacking it on top of the blue bowl.
    \item \textbf{Block-Cabinet} involves opening the cabinet, grasping a small red block from the table, and placing it in the cabinet.
    \item \textbf{Paper-Towel} involves precisely grasping a paper towel roll from the cabinet and placing it upright on a wooden stand on the table.
    \item \textbf{Hang-Towel (deformable object)} involves grasping a towel by the corner and hanging it on a hook attached to the cabinet.
    \item \textbf{Rice-Cooker (long horizon)} involves pouring a cup of rice into the inner pot of a rice cooker, and putting the inner pot in the rice cooker.
\end{itemize}
Each of these tasks involves positional and visual generalization, and requires reasonably precise robotic manipulation. We curate the finetuning datasets by teleoperating the robot and collecting a dataset of expert trajectories. 

We train all methods on the pretraining / co-training datasets for 100K steps and then finetune to the evaluation tasks (task-specific parameters shown in Table. \ref{tab:task-hyperparams}.) For cotraining experiments, we mix up the robot and video datasets and sample batches uniformly from the mixture dataset, where each batch may contain action-labeled and action-free data. We then apply the method-specific masking techniques and optimize the cotraining loss. For each task, we evaluate in scenarios approximately similar to those encountered during data collection (referred to as in-distribution), and we also construct an out-of-distribution evaluation setting by introducing distractions that are unseen in the finetuning dataset, as shown in Fig~\ref{fig:eval_setup}. To ensure statistically significant evaluation, we test each task on a fixed set of randomly chosen initialization positions. We provide details for the task-specific setups in Appendix \ref{app:task_set_up}.

\begin{figure}[t]
    \centering
    \includegraphics[width=\linewidth]{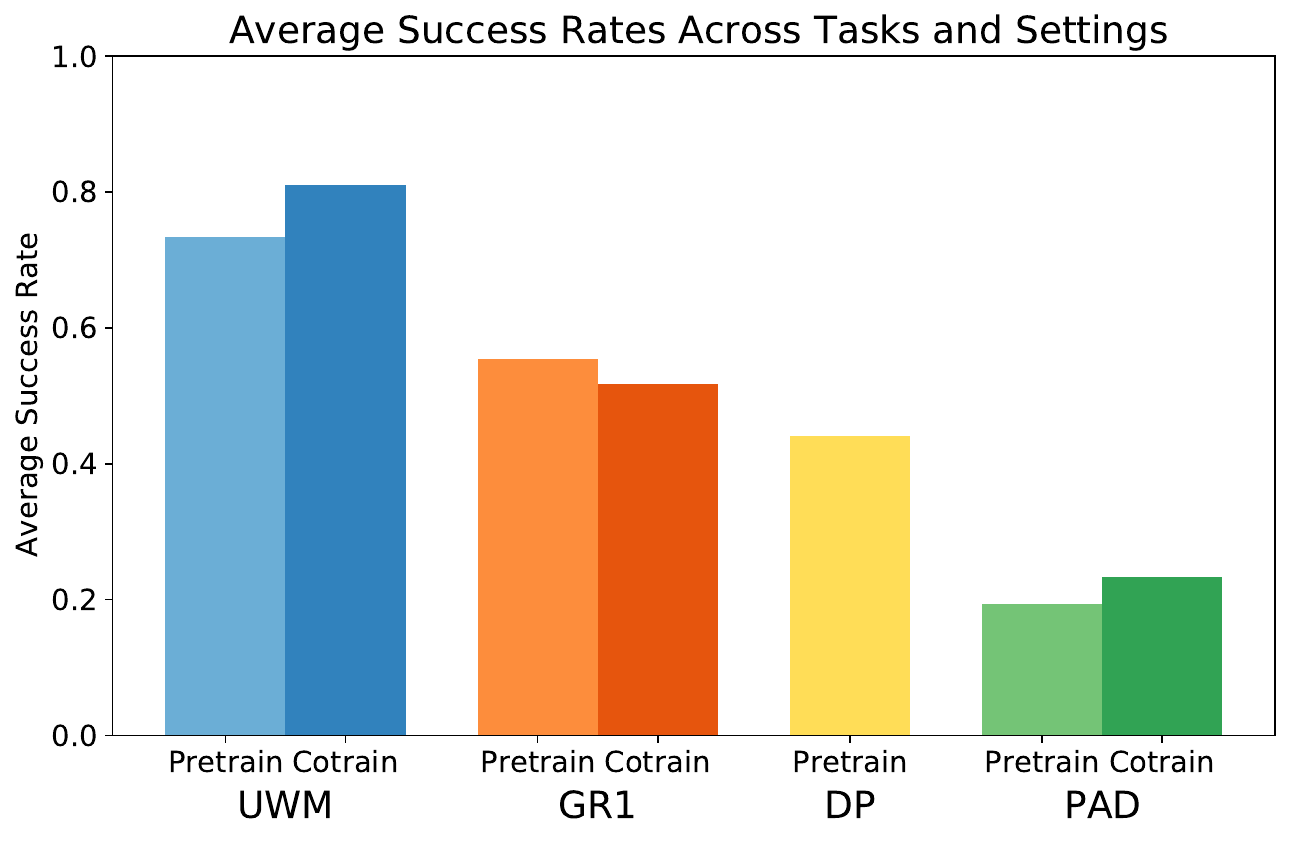} 
    \caption{\footnotesize{Average success rates across all real robot tasks and in-distribution and out-of-distribution settings. \algname{} exhibits strong performance and can further improve by co-training from action-free videos.}}
    \label{fig:exp-droid-bar}
\end{figure}

\subsubsection{Discussion}

\begin{table*}[t]
\centering
\caption{\footnotesize{Evaluation on LIBERO benchmark.}}
\begin{tabular}{lccccc|c}
\toprule
& Book-Caddy & Soup-Cheese & Bowl-Drawer & Moka-Moka & Mug-Mug & Average \\
\midrule
\algname{} \tiny{(Ours)}
  & \textbf{0.91 \std{0.07}}
  & \textbf{0.93 \std{0.01}}
  & \textbf{0.80 \std{0.02}}
  & \textbf{0.68 \std{0.02}}
  & \textbf{0.65 \std{0.01}}
  & \textbf{0.79 \std{0.11}} \\
DP
  & 0.73 \std{0.10}
  & 0.88 \std{0.02}
  & 0.77 \std{0.02}
  & 0.65 \std{0.03}
  & 0.53 \std{0.05}
  & 0.71 \std{0.12} \\
PAD
  & 0.78 \std{0.04}
  & 0.47 \std{0.04}
  & 0.74 \std{0.05}
  & 0.59 \std{0.08}
  & 0.25 \std{0.04}
  & 0.57 \std{0.19} \\
GR1
  & 0.77 \std{0.03}
  & 0.65 \std{0.05}
  & 0.62 \std{0.03}
  & 0.46 \std{0.04}
  & 0.38 \std{0.05}
  & 0.58 \std{0.14} \\
\bottomrule
\end{tabular}
\label{tab:exp-robomimic}
\end{table*}

We report the results on the real robot experiments in Table \ref{tab:combined_results} and the average performance in Figure \ref{fig:exp-droid-bar}. For each method an task, we provide results in the in-distribution (ID) and out-of-distribution (OOD) scenarios. Furthermore, for methods that support co-training on videos, we additionally report the results of co-trained models (separated by "/").

We first examine the pretraining results in the in-distribution setting. This set of experiments reflect the models' ability to accurately capture the expert policy's distribution. We find that \algname{} achieves the highest success rates across all five tasks among the methods, surpassing the best baseline by as much as 20\%. This demonstrates the strength of coupled action-video diffusion in absorbing rich dynamic information from multitask datasets. In particular, since the model is trained to capture all possible conditional and marginal distributions, it is instilled with an understanding of the causal relationship between actions and image observations, explaining its superior performance compared to joint prediction models such as GR1 and PAD. GR1 consistently outputs the second best results, establishing a strong baseline performance for deterministic regressive models. Diffusion Policies fail to leverage the rich and dynamic pixel information in the pretraining datasets, being inefficient at learning from diverse multitask trajectories. PAD achieves the lowest success across the board. We attribute its low performance largely to the conditioning via concatenation. Compared to \algname{} which takes in image features preprocessed by an encoder, PAD takes in raw pixels, thus needing to incorporate the feature extraction in the same transformer model. This limits its performance at accurately capturing the conditional action distribution without expanding model capacity.

We then examine the OOD scenarios. This set of experiments tests the models' robustness to distribution shifts. We find that all models experience performance drops in the presence of visual distractions. This is especially pronounced in Stack-Bowls, Block-Cabinet, and Hang-Towel. In the Paper-Towel task, the models seem unaffected by the visual distractions, potentially due to the task not requiring the models to pay attention to the table top when grasping the paper towel. Despite a slight performance drop compared to the ID setting, we find \algname{} to outperform the baselines, showcasing strong robustness under distribution shifts.

Finally, we test the methods' potential to scale with videos by cotraining with action-free videos. Results are reported after the / in each entry of Table \ref{tab:combined_results}. We find \algname{} to consistently improve performance when exposed to additional videos during pretraining. This suggests using diffusion time steps for masking as an effective strategy for co-training on multimodal data. While GR1 is able to learn from videos by masking the actions with a learnable token, we found mixed results of the cotrained model. In Stack-Bowls, Paper-Towel, and Rice-Cooker, the cotrained GR1 model is worse than the pretrained model, which implies that incorporating videos dilutes the action learning signal. While PAD showcases weak positive transfer as a result of cotraining, its baseline performance is suboptimal. In Table. \ref{tab:exp-ood}, we perform evaluations in a larger set of OOD scenarios and found video cotraining to provide significant gains in those settings.

\subsection{Simulated Experiments}
\label{sec:exp-sim}

\begin{figure}[t]
    \centering
    \includegraphics[width=\linewidth]{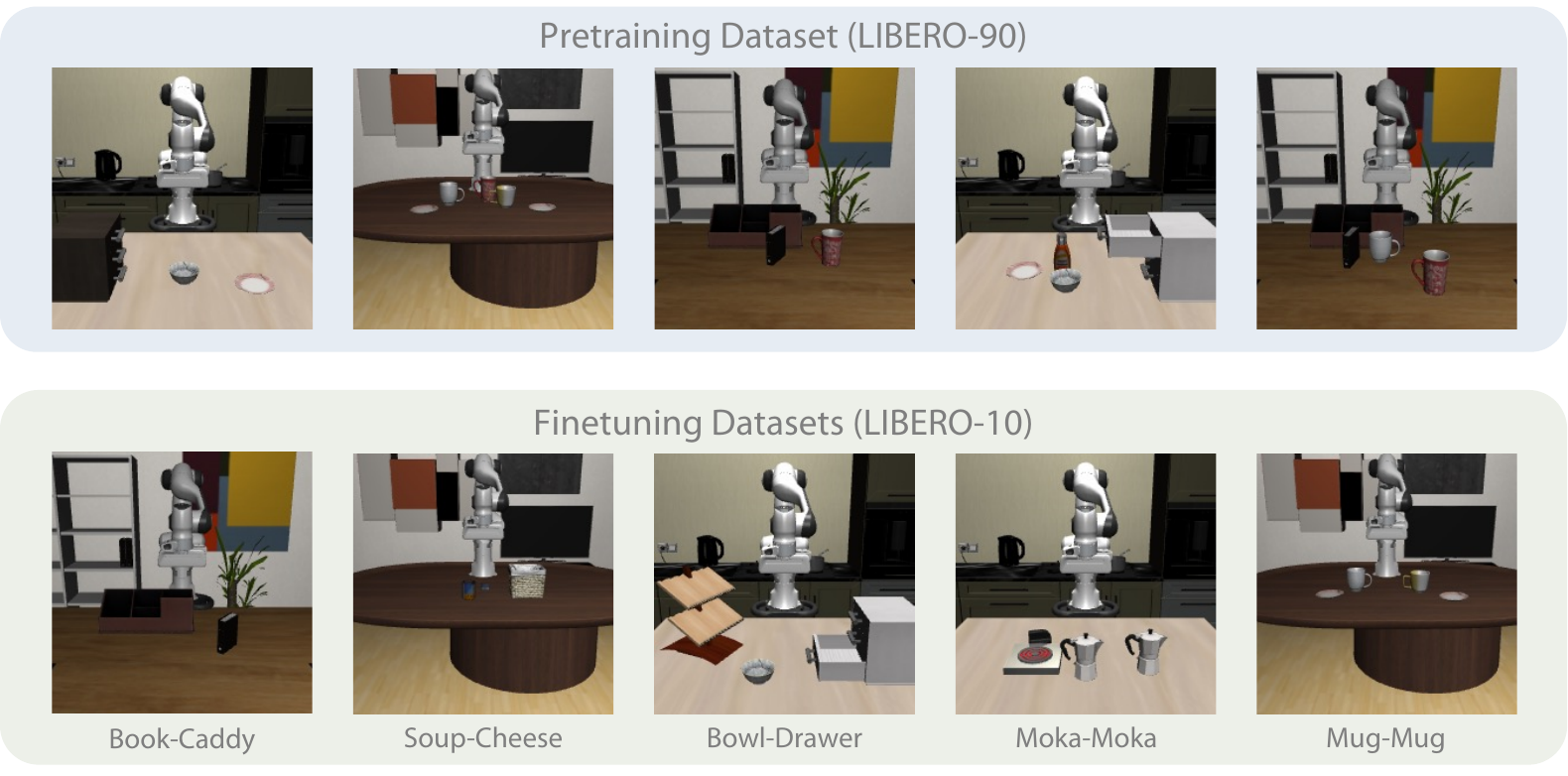} 
    \caption{Visualization of the LIBERO datasets. The pretraining dataset (LIBERO-90) consists of 90 tasks sampled across the kitchen, living room, and study scenes. The finetuning datasets (LIBERO-10) consist of 10 tasks used for evaluation. Tasks from LIBERO-10 are fine-tuned and evaluated under distribution shifts, with unseen initializations and modified configurations.}
    \label{fig:env-libero}
\end{figure}

To validate these findings in standard community benchmark settings, we evaluate the methods on the LIBERO~\cite{liu2023libero} simulation benchmark. The LIBERO-100 benchmark consists of 90 training environments across multiple scenes and 10 evaluation environments, each with accompanying expert demonstrations. We combine the demonstrations from the 90 training environments to construct a multitask training dataset, and finetune on a random subset of the evaluation environments, shown in Fig \ref{fig:env-libero}. To evaluate the methods' generalization capabilities, we introduce distribution shifts to evaluation environments by enlarging the range of initialization for all objects and removing objects from the scene. The details for this setup is described in Appendix \ref{app:sim-exps}.

We pretrain each method on the multitask dataset for 100K gradient steps, and finetune on the downstream tasks for 10K gradient steps. We finetune 3 random seeds for each method on each environment, and evaluate on 50 different initializations. Table \ref{tab:exp-robomimic} reports the average success rates across initializations with confidence intervals across random seeds. \algname{} achieves the highest success rates across the evauation tasks in the out-of-distribution setting. DP achieves the second highest performance, followed by GR1 and PAD. These results imply that \algname{} effectively learns from large robotic datasets, due to its use of pixel reconstruction as an auxiliary signal and the independent diffusion timesteps instilling the model with a causal understanding of actions and observations.

Although our method showed an improvement over baselines, we note that the improvement in OOD scenarios is less than the real world experiments. We hypothesize this to be an artifact of current simulations having simpler dynamics than what we see in the real world. 

\subsection{Analysis and Ablation Experiments}

\label{sec:exp-ablation}
In this section, we conduct analysis and ablation experiments to help understand the various components and design choices in \algname{}. We provide additional experiments in Appendix. \ref{app:additional-exps}.

\subsubsection{Forward Dynamics}

\begin{figure}[t]
    \centering
    \includegraphics[width=\linewidth]{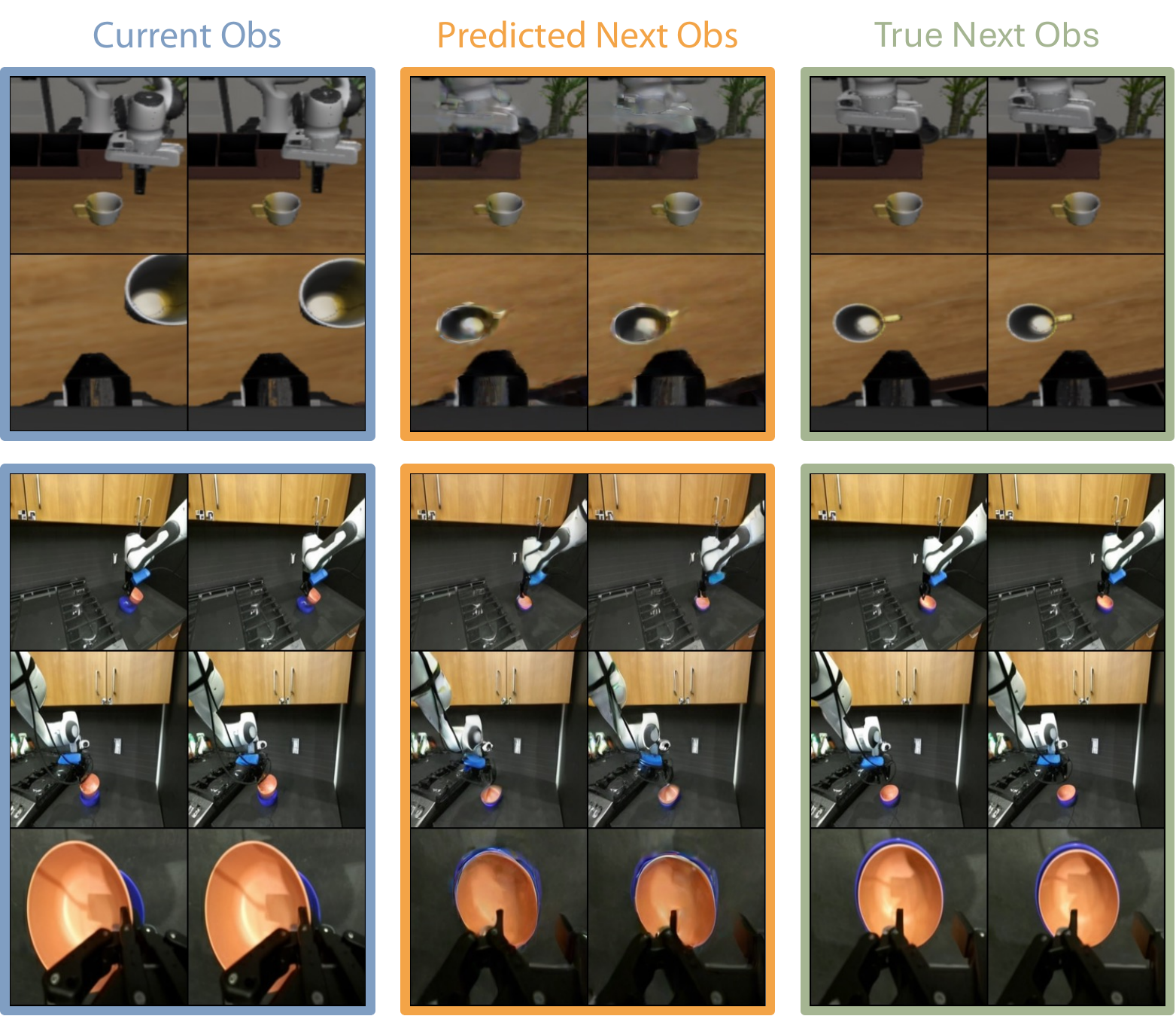} 
    \caption{\footnotesize{Visualization of the forward dynamics predictions. The model accurately predicts the robot and object poses conditioned on the initial observation and actions.}}
    \label{fig:viz-forward}
\end{figure}

To examine the world modeling component of \algname{}, we visualize the forward dynamics prediction of \algname{} on simulated and real-world domains. To generate samples from the forward dynamics model, we perform image diffusion while fixing the action diffusion timestep to 0 and setting the action tokens to be the ground truth actions. As shown in Fig. \ref{fig:viz-forward}, \algname{} accurately predicts the image observations conditioned on actions, closely resembling the ground truth image observations. This implies that \algname{} can effectively model the conditional distribution. 

\subsubsection{Inverse Dynamics} 

We evaluate the inverse dynamics mode of \algname{} on trajectory tracking, where we provide a reference expert trajectory and query the inverse dynamics model to track it. Specifically, for each reference trajectory, we reset the simulation environment to match the exact initial state of this trajectory. At each step, we take the ground truth future observations from the trajectory and use the inverse dynamics mode of a finetuned \algname{} to generate corresponding actions. Table~\ref{tab:exp-inverse-dynamics} shows the results of tracking 50 trajectories from the LIBERO training datasets. We find that given the same time limit as the trajectory length, the inverse dynamics model achieves a higher success rate than the policy. This implies that actions generated by the inverse dynamics adhere more closely to the reference trajectory. We note that while the policies deviate from the reference trajectories, they eventually recover and solve the tasks given enough time. 

\begin{table}[!t]
\centering
\caption{Trajectory Tracking Experiments}
\begin{tabular}{lcc}
\toprule
& Book-Caddy & Soup-Cheese \\
\midrule
Policy (1000 steps) & 1.00 \std{0.00} & 0.97 \std{0.01} \\
Policy (trajectory length) & 0.47 \std{0.02} & 0.26 \std{0.02} \\
Inverse dynamics (trajectory length) & 0.65 \std{0.01} & 0.55 \std{0.02} \\ 
\bottomrule
\end{tabular}
\label{tab:exp-inverse-dynamics}
\end{table}

\subsubsection{Categorized OOD Experiments}

\begin{figure}[t]
  \centering
  \includegraphics[width=1.0\linewidth]{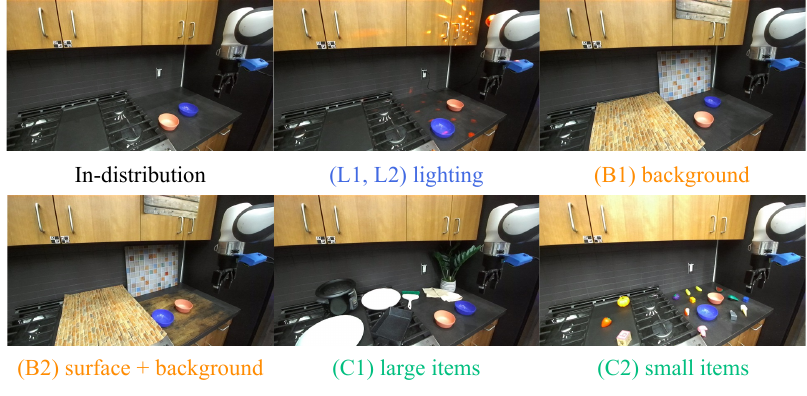}
  \caption{Visualization of categorized out-of-distribution (OOD) settings. We construct scenes with varied lighting conditions, backgrounds, and clutter to analyze the models' generalization patterns.}
  \label{fig:vis_ood}
\end{figure}

We evaluate \algname{} and DP in several more out-of-distribution (OOD) settings to study their generalization patterns. As shown in Fig. \ref{fig:vis_ood}, we construct scenes with varied lighting conditions (including static and Disco lights), backgrounds, and clutter. For each scene, we randomly select 5 initializations to evaluate. Results in Table. \ref{tab:exp-ood} show that across the board, \algname{} cotrained on videos (co) is significantly more robust than both \algname{} (pre) and DP pretrained on robot data. 

\definecolor{lighting}{HTML}{4169E2}
\definecolor{background}{HTML}{FF8C02}
\definecolor{clutter}{HTML}{00BF7F}
\begin{table}[t]
\centering
\caption{OOD Performance on Stack-Bowls and Block-Cabinet}
\begin{tabular}{lccc|ccc}
\toprule
 & \multicolumn{3}{c|}{Stack-Bowls} & \multicolumn{3}{c}{Block-Cabinet} \\
 & \algname{} \tiny{(Co)} & \algname{} \tiny{(Pre)} & DP & \algname{} \tiny{(Co)} & \algname{} \tiny{(Pre)} & DP \\
\midrule
{\color{lighting} L1} & 4/5 & 4/5 & 2/5 & 5/5 & 5/5 & 3/5 \\
{\color{lighting} L2} & 3/5 & 2/5 & 2/5 & 4/5 & 0/5 & 0/5 \\
{\color{background} B1} & 4/5 & 3/5 & 3/5 & 4/5 & 3/5 & 2/5 \\
{\color{background} B2} & 3/5 & 1/5 & 2/5 & 1/5 & 0/5 & 0/5 \\
{\color{clutter} C1} & 3/5 & 2/5 & 2/5 & 0/5 & 0/5 & 0/5 \\
{\color{clutter} C2} & 4/5 & 3/5 & 1/5 & 1/5 & 0/5 & 0/5 \\
\midrule
\textbf{All} & \textbf{21/30} & \textbf{15/30} & \textbf{12/30} & \textbf{15/30} & \textbf{8/30} & \textbf{6/30} \\
\bottomrule
\end{tabular}
\label{tab:exp-ood}
\end{table}

\subsubsection{Real-World Learning from Scratch}
\begin{figure}[t]
  \centering
  \includegraphics[width=1.0\linewidth]{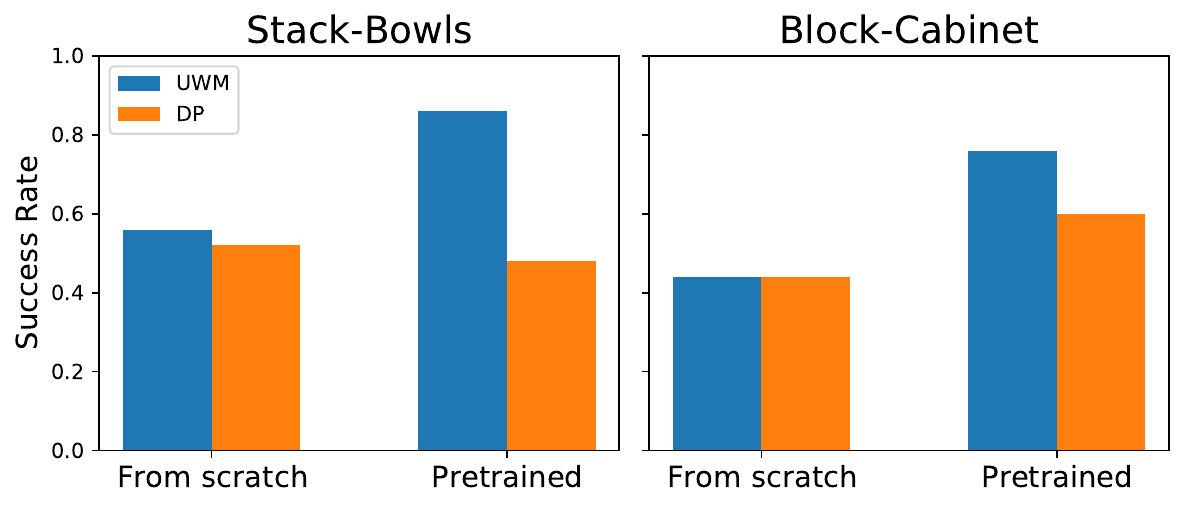}
  \caption{Training models from scratch vs finetuning pretrained models. \algname{} scales more effectively with pretraining than DP.}
  \label{fig:scratch_vs_finetuning}
\end{figure}

To study \algname{}'s ability to scale with pretraining, we train \algname{} and DP on the task-specific expert demonstrations \textit{from scratch} for the same number of steps as the finetuning stage of the experiments in Table. \ref{tab:combined_results}. As shown in Fig. \ref{fig:scratch_vs_finetuning}, we find that \algname{} and DP perform similarly when trained from scratch. However, \algname{} scales from pretraining more effectively than DP. 
\section{Related Work}

\paragraph{\textbf{Imitation Learning}} Imitation learning (IL) for robotics is a paradigm in which robots learn to perform tasks by learning behaviors from experts, typically via teleoperation. A common approach within the imitation learning family is behavior cloning, where supervised learning techniques are applied to replicate expert actions from the provided demonstrations. In particular, these methods are useful for tasks with well defined inputs such as manipulation. 

One common challenge for problems cast in the BC framework is the inability to fit multi-modal action distributions \cite{bagnell}. Previous methods have attempted to solve this by attempting to fit multiple pre-defined distributions \cite{robomimic2021}, using architectures amenable to modeling high-dimensional distributions \cite{bet,vqbet,act, zhao2024alohaunleashedsimplerecipe}, and more recently, generative models such as diffusion \cite{chi2023diffusionpolicy}. Diffusion models, in particular, have shown to scale favorably to both a large number of demonstrations ~\cite{octo, pi0}, and dexterous behaviors \cite{ditpolicy}. Although the diffusion framework has shown the ability to scale, at their core, these formulations rely on access to \textit{high quality} action data. Despite recent efforts from the community to open source large amounts of data \cite{droid, embodimentcollaboration2024openxembodimentroboticlearning}, the magnitudes of readily available data pales in comparison to the internet scale data that is used to train state of the art foundation models such as LLMs (Large Language Models) and VLMs (Vision Language Models). Alternative formulations to scaling robotic policies focus on leveraging pre-trained foundation models in order to leverage their common-sense reasoning \cite{kim24openvla, wen2024tinyvlafastdataefficientvisionlanguageaction} using autoregressive techniques. Although these efforts are promising, they are still heavily reliant on access to high quality action data and focus on increasing generalization. 

\paragraph{\textbf{Learning from Videos}}In order to scale large robot foundation models, an appealing approach lies in leveraging video as a source of abundant data. Video data, however, does not contain explicit actions and may contain a significant cross-embodiment gap. In order to address these issues, hand-engineered solutions are often used in order to extract semantic information and map this information to the physical robot. For example, \cite{wen2024anypointtrajectorymodelingpolicy, xiong2021learningwatchingphysicalimitation} both use keypoints to map actions from video models to the robots themselves. Alternative methods use predicted future points and maps these to rigid body transforms explicitly in order to transfer from internet trained videos to robots \cite{bharadhwaj2024track2actpredictingpointtracks}. Other work often explicitly track human hand trajectories and and contact patches in order to leverage data from human videos ~\cite{wang2023mimicplaylonghorizonimitationlearning, bahl2023affordanceshumanvideosversatile}. 
 
An alternative approach to leveraging video data relies on large scale pre-training on robot video datasets. For example, \cite{gr-1,lapa} use an autoregressive style prediction to pre-train a video and language model which is then fine-tuned on robot actions in a second stage. Other approaches use diffusion models in order to predict and supervise on dense future frames combined with an action diffusion transformer \cite{vpp}. These use a two-stage process that relies on fine-tuning pre-trained vision models that may not contain robot information. Most importantly, by using a decoupled architecture, they limit the feature sharing capabilities between the video and action data. Closest in spirit to our approach is PAD~\cite{pad} which trains a joint video-action model using diffusion as its core mechanism. Their approach, however, uses a \textbf{shared} diffusion time step between all the modalities which we hypothesize leads to a sub-optimal shared representation that lacks a causal understanding between the underlying video and action models. We show that by having independent diffusion timesteps, our policy performs better in both in distribution and out of distribution scenarios. 

\paragraph{\textbf{Unified Inference}} Unified multi-modal models for both decision making and general inference have recently become an emerging topic due to the potential of feature sharing between modalities. \cite{unimask} explores this topic from the perspective of decision making and shows that masking tokens is an effective way to share information across the decision making process itself. \cite{unidiffuser} studies this problem from the diffusion perspective on image and text generation. Their results show that by having flexible control of each modality, and thus controlling the marginal, conditional, and joint probabilities, the model is able to do share features and show an increased performance for each individual modality. Our framework builds upon the core insight from this work and studies this from the perspective of joint video and action modeling. 

Finally, recent efforts exist in order to combine advantages from both autoregressive and more continuous approaches. \cite{transfusion} combines the ability to do both autoregressive language generation and diffusion based image generation in one framework. Their framework shows efficient scalability and feature sharing. \cite{chen2024diffusionforcingnexttokenprediction} also provides an alternative way to bridge the gap between autoregressive and diffusion techniques. Although the framework provides a mechanism for doing flexible inference that combines the capabilities of continuous and discrete approaches, the multi-modal feature sharing capabilities have not been shown yet. 
\section{Discussion}
In this work, we present \algnamefulls{}, a diffusion based framework that unifies policy learning and world modeling into a single flexible framework. We instantiate \algname{} with a coupled conditional diffusion process using separate timesteps for actions and future observations. During training, the model is exposed to all combinations of timesteps covering various conditional and marginal distributions, instilling the model with an understanding of the causal relationship between actions and future observations. This distinguishes \algname{} from traditional imitation learning approaches, which often lack a nuanced understanding of causal dependencies. Moreover, the independent diffusion timesteps allow for a natural connection between noising and partial masking, enabling the use of action-free videos for co-training, as well as for marginalization and conditioning of the variables by appropriately setting timesteps. The resulting model is then able to flexibly perform inference as a policy, a video prediction model, a forward dynamics model, and an inverse dynamics model. We show through a thorough experimental evaluation that \algname{} provides significant gains over imitation learning across the board by enhancing large scale pretraining from robotic datasets.

\section{Limitations}
While \algname{} shows promising results, there are several avenues for future investigation. Firstly, the proposed model does not yet learn from large scale human videos, bridging the embodiment gap. Additionally, while \algname{} shows an improvement on action prediction, the forward dynamics reconstruction may often contain artifacts which may reduce efficacy when planning with this model. We believe this can be addressed by incorporating the latest progress in the generative model literature. Finally, we expect to see further improvement by leveraging denser video prediction. 


\section*{Acknowledgements}
We thank members of the WEIRD lab at UW and the Toyota Research Institute for thoughtful discussions during the course of this work. CZ was supported by funding from the Toyota Research Institute and NSF Grant No. 2212310 during the course of this work. 
\bibliographystyle{plainnat}
\bibliography{references}

\clearpage
\appendix

\subsection{Additional Implementation Details} 
\label{app:implementation}

\subsubsection{Model Architecture}
We base our implementation of \algname{} on the diffusion transformer architecture with AdaLN conditioning \cite{Peebles2022DiT}. The inputs to the model are $(o, a_{t_a}, o'_{t_{o'}}, t_a, t_{o'})$, where $o := \{o^{i}_{0:h_o}\}_{i=1}^{n_c}$ is a sequence of observations from $n_c$ camera views, $a_{t_a} := a_{h_o:h_o + h_a}$ is a sequence of noisy actions, $o'_{t_{o'}}:= \{o^{i}_{h_o+h_a:2h_o + h_a}\}_{i=1}^{n_c}$ is a sequence of noisy observations from each camera view \textit{after} the actions, and $t_a, t_o'$ are diffusion timesteps. The current observations are encoded into features $\{\phi^i_{0:h_o}\}_{i=1}^{n_c}$ using a ResNet-18 \cite{He2015DeepRL} encoder, which is initialized the using ImageNet \cite{imagenet} pretrained weights and updated throughout training. The timesteps $t_a$ and $t_{o'}$ are encoded into features $\psi_a$, $\psi_{o'}$ via a sinusoidal embedding network \cite{ho2020denoising}. The image features are flattened and concatenated with the timestep embeddings and used to condition each transformer block via AdaLN layers \cite{Peebles2022DiT}.

The input sequence to the transformer consists of encoded tokens from $a_{t_a}$, $o'_{t_{o'}}$ and additional register tokens $r_{1:N_r}$. The actions are encoded to tokens using a shallow MLP encoder share across timesteps. For images, we follow the latent diffusion paradigm \cite{rombach2021highresolution} and downsample the raw image observations into latent space using a frozen VAE from Stable Diffusion XL \cite{podell2024sdxl}. The image latents are patchified into patch embeddings using a 3D convolution layer. We concatenate the action embeddings, the image patch embeddings, and the learnable register tokens along the sequence dimension and pass them as input to the transformer model. We add a learnable positional embedding to the inputs to encode positional information. To decode action and image noise predictions from the model outputs, we take the respective tokens (discarding registers) and decode then using shallow MLP networks. Note that the image noise predictions are in the latent space, and we only decode the final image prediction at the end of the sampling procedure.

\subsubsection{Training and Inference Details}

Given a transition tuple $(o, a, o')$ from sampled from the dataset, we first apply random cropping and augmentations to the image observations. The cropping and augmentation parameters are kept temporally consistent across $o$ and $o'$ but differ from camera view to camera view. We then sample action and observation diffusion timesteps $t_a, t_o'$ \textit{independently} from the uniform distribution $\mathcal U[0, T)$. These are used to sample noisy actions $a_{t_a}$ and observations $o'_{t_o'}$ according to the forward diffusion process. The tuple $(o, a_{t_a}, o'_{t_{o'}}, t_a, t_{o'})$ is passed as input to the model, which outputs the action and observation noise predictions $\epsilon_a, \epsilon_{o'}$. We train the model by optimizing the diffusion loss outlined in Eq. \ref{eq:objective}.

To co-train the model on video data, we combine a robot dataset and a video dataset to get a mixture dataset and sample batches of transition tuples from the mixture dataset uniformly at random. Each batch contains a mixture of video data and action data. For the action-free video samples in each batch, we manually set the corresponding action diffusion timesteps to $t_a = T$, and impute the missing actions with random actions drawn from the unit Gaussian distribution. The action diffusion loss is computed across all samples in a batch (both robot samples and video samples).

We optimize the model using the AdamW \cite{loshchilov2018decoupled} optimizer. For pretraining experiments, we use a constant learning rate. For finetuning experiments, we use a cosine annealed learning rate with warmup. We sample from the reverse diffusion processes using the DDIM \cite{song2021denoising} sampler to speed up inference. At deployment, we execute the the first $h'_a$ action predictions and replan. We provide all model and training hyperparameters in Table \ref{tab:hyperparams}.

\textbf{Tips for Tuning \algname{}} While \algname{} is generally stable with respect to hyperparameters, we find that for pretraining on highly multimodal datasets, increasing the number of registers helps improve performance. For new datasets, we recommend trying the default hyperparameters first and then tuning the number of registers for potential performance gains.

\begin{table}[t]
    \centering
    \caption{Hyperparameters}
    \begin{tabular}{ll}
        \toprule
        \textbf{Parameter} & \textbf{Value} \\
        \midrule
        \textbf{Model} &  \\
        Observation Length $h_o$ & 2 \\
        Observation Encoder & ResNet-18 \\
        Image VAE & SDXL \\
        Image Shape & [224, 224, 3] \\
        Latent Image Shape & [28, 28, 4] \\
        Patch Shape & [4, 4, 2] \\
        Action Length $h_a$ & 16 \\
        Rollout Length $h'_a$ & 8 \\
        Embed Dim & 768 \\
        Timestep Embed Dim & 512 \\
        Depth & 12 \\
        Num Heads & 12 \\
        MLP Ratio & 4 \\
        QKV Bias & True \\
        Num Registers $N_r$ & 8 \\
        \midrule
        \textbf{Diffusion} \\
        Beta Schedule & squaredcos\_cap\_v2 \\
        Num Training Diffusion Steps & 100 \\
        Num Inference  Diffusion Steps & 10 \\
        Sampler & DDIM \\
        Clip Sample & True \\
        \midrule
        \textbf{Training} &  \\
        Batch Size (Distributed) & 36 $\times$ 4 (pretraining) \\
        &  36 $\times$ 2 (finetuning) \\
        Optimizer & AdamW \\
        Learning Rate & $1e^{-4}$ \\
        Weight Decay & $1e^{-6}$ \\
        Betas & [0.9, 0.999] \\
        Epsilon & $1e^{-8}$ \\
        LR Schedule & constant (pretraining)\\
        & cosine w/ warmup (finetuning) \\
        LR Warmup Steps & 1000 \\
        Action Loss Weight $w_a$ & 1.0 \\
        Image Loss Weight $w_{o'}$ & 1.0 \\
        \bottomrule
    \end{tabular}
    \label{tab:hyperparams}
\end{table}

\subsubsection{Training Compute}
Training a \algname{} on the DROID dataset for 100K gradient steps with the hyperparameters shown in Table \ref{tab:hyperparams} takes 24 hours on 4 NVIDIA A100 GPUs using Pytorch DDP.

\subsection{Baseline Details}
\subsubsection{Diffusion Policies} We base our implementation of diffusion policies on the \algname{} model. We remove the image tokens, image diffusion timestep, and registers and keep everything else identical. This is equivalent to the Transformer version of the original diffusion policy~\cite{chi2023diffusionpolicy} and similar to the architecture in \cite{ditpolicy}.
\subsubsection{PAD} We base our implementation of PAD on the \algname{} model, replacing coupled action-image diffusion with joint diffusion, and condition the model by concatenating the clean current observations to the noisy future observation predictions along the channel dimension. The diffusion timestep is still passed into the transformer via AdaLN. While the original PAD~\cite{pad} method predicts consecutive actions and future frames, we adapt it to predict sequences of actions and the following observations (same as \algname{}) to isolate the effect of key design differences such as joint video-action diffusion and conditioning method.
\subsubsection{GR1} We use a custom implementation of the GR1 model adapted to have the same input-output format as \algname{}. Instead of regressing consecutive actions and observations, we predict a sequence of actions and the following image observations. GR1 conditions on the current observations by passing the ViT encoded observation tokens through a Perceiver resampler from Flamingo \cite{flamingo}, and then concatenating the resulting tokens to the input sequence of the transformer model. The rest of input sequence for the transformer consists of learnable action and observation tokens. The output tokens are passed into respective decoders (MLP for actions, DiT decoder for image patches) to regress the modalities. 

\subsection{Additional Details on Real-World Experiments}
\label{app:real-exps}

\subsubsection{Robot Setup}
\label{app:task_set_up}

\begin{figure}[t]
    \centering
    \includegraphics[width=0.8\linewidth]{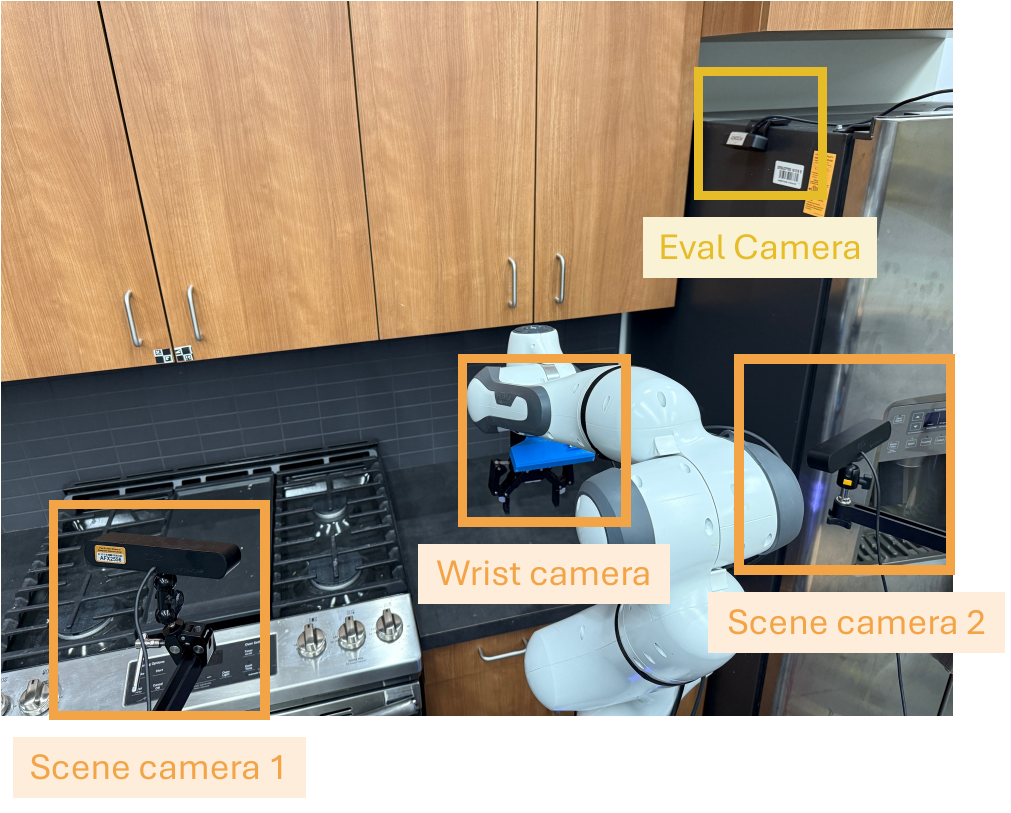} 
    \caption{Setup of the robot experiments. We adopt the DROID \cite{droid} setup which consists of two scene cameras and one wrist camera. We use an additional evaluation camera to track the initialization of evaluation seeds.}
    \label{fig:robot-setup}
\end{figure}

We conduct real-world experiments using a Franka Panda robot in the DROID \cite{droid} setup. As shown in Fig. \ref{fig:robot-setup} the robot's observation space consists of two scene cameras and a wrist camera (visualized in Fig. \ref{fig:robot_view}. We additionally mount an overhead camera to track the initializations during evaluation. The robot operates at a control frequency of 10 Hz, allowing us to have responsive and smooth task execution. The action space is defined by a delta end-effector (EE) pose, which specifies incremental positional and rotational adjustments relative to the current pose. Additionally, the gripper state is represented using a single continuous dimension, where 0 indicates the gripper is open and 1 indicates the gripper is closed.

\begin{table}[t]
    \centering
    \caption{Task-Specific Parameters}
    \begin{tabular}{lccc}
        \toprule
        & \# demos & \# finetuning steps & \# eval conditions \\
        \midrule
        Stack-Bowls & 50 & 10K & 50 \\
        Block-Cabinet & 50 & 10K & 50 \\
        Paper-Towel & 100 & 20K & 50 \\
        Hang-Towel & 50 & 10K & 50 \\
        Rice-Cooker & 150 & 50K & 20 \\
        \bottomrule
    \end{tabular}
    \label{tab:task-hyperparams}
\end{table}

\subsubsection{Tasks}

We provide a detailed description of each real-world task shown in Fig. \ref{fig:eval_setup} and the task-specific settings in Table \ref{tab:task-hyperparams}.

\begin{itemize}
\item \textbf{Stack-Bowls:} the robot needs to pick up the red bowl on the counter and place it in the blue bowl. The positions of the bowls are randomized across the counter top. A rollout is successful if the red bowl is placed securely inside the blue bowl. For the OOD setup, we open the top cabinet, the bottom drawer, and place unseen objects on the counter and stovetop.

\item \textbf{Block-Cabinet:} the robot needs to (1) open the left cabinet door by grasping the handle, and (2) pick up the red block from the counter top and place it on the bottom level of the cabinet. The position of the red block is randomized across the counter. A rollout is successful if the block is placed securely in the cabinet. For the OOD setup, we open the bottom drawer and place unseen objects on the counter and stovetop.

\item \textbf{Paper-Towel:} the robot is tasked to take out a paper towel placed in the open cabinet and place it vertically on a base plate on the counter. The position of the paper towel is randomized across the cabinet shelf, and position of the base plate is randomized across the counter top. Success is counted if the paper towel is placed securely on the base plate and does not topple. For the OOD setup, we open the bottom drawer and place unseen objects on the counter and stovetop. 

\item \textbf{Hang-Towel: } the robot is tasked to pick up a towel from the counter and hang it on a hook on the cabinet. The position and shape of the towel are randomized during data collection. For evaluation, we fold the towel carefully to ensure standardization (Fig. \ref{fig:eval-tracker}). A rollout is successful if the towel hangs on the hook and does not slip off. For the OOD setup, we open the bottom drawer and place unseen objects on the counter and stovetop.

\item \textbf{Rice-Cooker: } this is a multistage task that involves (1) picking up a cup of rice, (2) pouring the rice into the bowl, (3) placing the cup back on the counter, (4) picking up the bowl and placing it in the rice cooker. The positions of all objects are randomized. A rollout is successful if there is minimal spill of rice and the bowl is placed securely in the rice cooker. We find this task to be particularly challenging and hence only evaluate on 20 initializations that are close to the dataset distribution. We do not evaluate this task in OOD settings.

\end{itemize}

\subsubsection{Evaluation Protocol}
\begin{figure}[t]
    \centering
    \includegraphics[width=\linewidth]{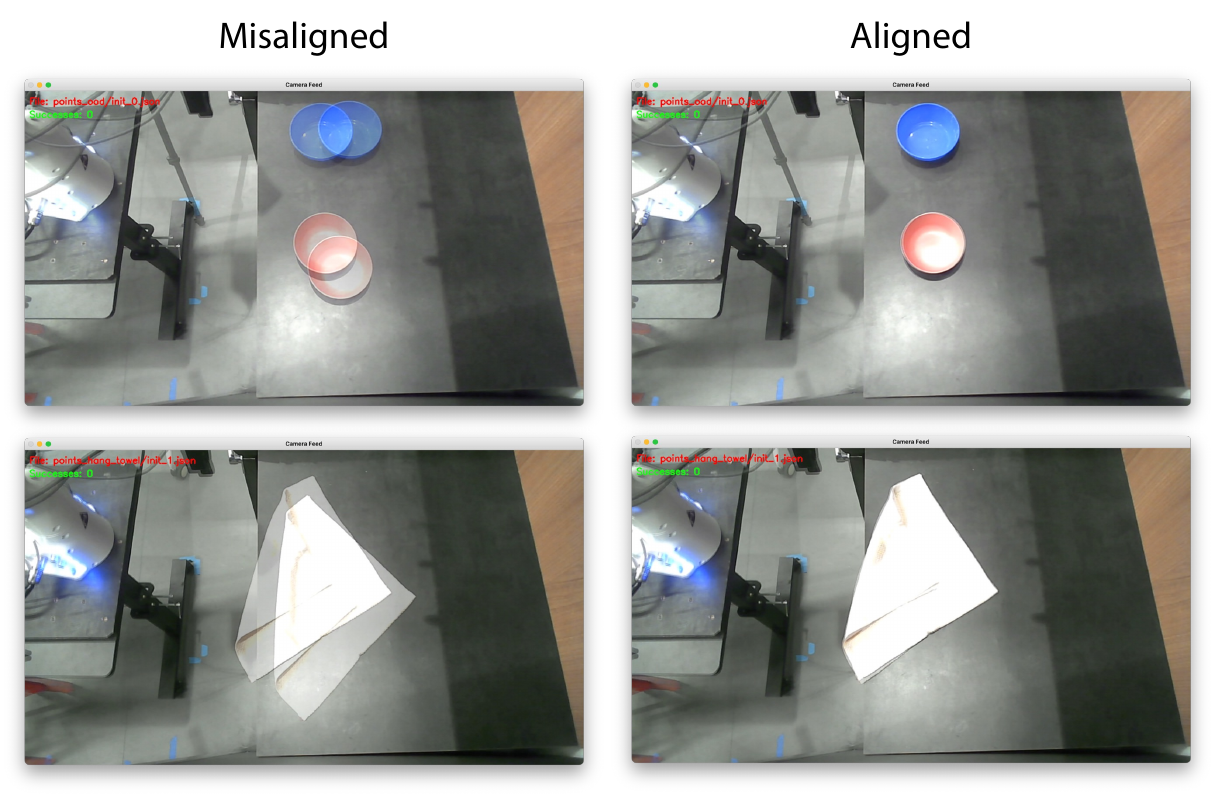} 
    \caption{Screenshots of the evaluation tracker. The tracker overlays the reference initial frame to the current frame. The same interface is used to tracker the initialization for all real robot tasks.}
    \label{fig:eval-tracker}
\end{figure}

To ensure fairness of real-robot evaluations, we use an overhead camera and a Python program to systematically track randomizations. As shown in Fig. \ref{fig:eval-tracker}, the program overlays the reference frame onto the current frame, so the user can adjust the objects to match the reference frame.  All tasks except Rice-Cooker are evaluated on 50 randomly generated configurations. We find Rice-Cooker particularly challengeing and evaluate on 20 configurations close to the data distribution. To mitigate the effects of camera shake due to the mounting mechanism, each method is given three attempts per initialization, making for a more robust evaluation across trials.

\subsubsection{Failure Modes}

\begin{figure}[t]
    \centering
    \includegraphics[width=\linewidth]{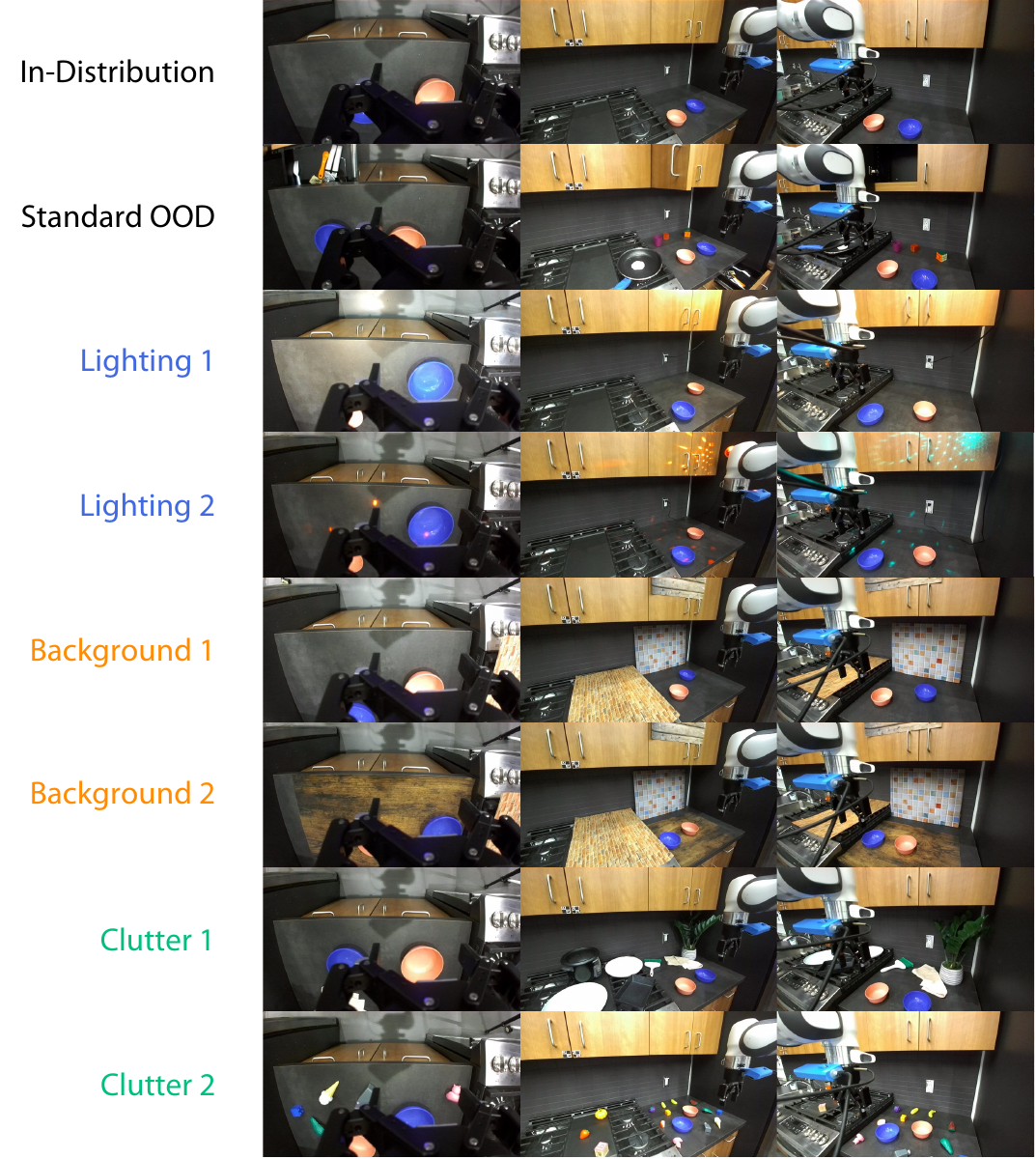} 
    \caption{Visualization of the robot's perspective in in-distribution, standard out-of-distribution (Table. \ref{tab:combined_results}), and categorized out-of-distribution (Table. \ref{tab:exp-ood}) scenarios.}
    \label{fig:robot_view}
    \vspace{-0.3cm}
\end{figure}

We provide a description of some common failure modes in the real-world experiments. Although we utilized three cameras to maximize coverage (Fig. \ref{fig:robot_view}), certain angles resulted in objects being visible to only one camera. These limited viewpoints made some initializations more challenging for the robot to complete the tasks successfully. Additionally, variability in object behavior contributed to task failures. For instance, in the Paper-Towel task, the robot often places the paper towel on the wooden platform, but the angle of placement may cause the paper towel to topple over. In the Stack-Bowls task, a source of failure for baseline methods is their inability in distinguishing between the blue bowl and the distractor when attempting to locate the blue bowl after picking up the pink bowl. This issue does not occur with our proposed method.

\section{Additional Details on Simulated Experiments}
\label{app:sim-exps}

\subsubsection{Simulated Environments}
LIBERO \cite{liu2023libero} is a simulated robotic benchmark designed to evaluate lifelong learning algorithms. It involves controlling a 7-DoF Franka Panda robot to complete various tasks across different scenes. The LIBERO-100 benchmark consists of 100 tasks distributed across three scenes (kitchen, living room, study), each with 50 accompanying expert demonstrations. The 100 tasks are split into 90 tasks for training (LIBERO-90) and 10 tasks for evaluation (LIBERO-10)

For our experiments, we use the combined LIBERO-90 dataset as the pretraining data, totaling 4500 trajectories. We evaluate on a random subset of 5 tasks from LIBERO-10. For each task, we finetune the pretrained models on 50 expert demonstrations. To evaluate the generalization capabilities of the methods, we modify the simulation configuration to introduce distribution shifts during the evaluation. Specifically, we increased the initialization range of each object by $0.03$ to generate unseen initializations and removed background objects to introduce visual distribution shifts. Visualizations of the evaluation environments are shown in Fig \ref{fig:env-libero}, and we provide a description of the evaluation tasks below.
\begin{enumerate}
\item Book-Caddy: the robot needs to pick up the book from the table top and place it in the back of a caddy.
\item Soup-Cheese: the robot needs to place the alphabet soup and the cheese in the basket in sequence.
\item Bowl-Drawer: the robot needs to pick up the bowl, place it in the bottom drawer, and close the drawer.
\item Moka-Moka: the robot needs to pick up the two Moka cups from the table and place them on the electric stove.
\item Mug-Mug: the robot needs to place the left mug in the left plate and place the right mug in the right plate.
\end{enumerate}
\subsection{Additional Experiments}
\label{app:additional-exps}

\subsubsection{Ablations of Design Choices}
To understand the effect of \algname{}'s design choices, we conduct ablation studies on two simulated tasks from the LIBERO environment. Specifically, we want to (1) understand the effect of registers on task performance, and (2) compare the use of AdaLN for observation conditioning with cross attention \cite{dosovitskiy2021an}. For each model, we train them on the single-task datasets from scratch (without pretraining), and evaluate on 50 initializations across 3 seeds.

\begin{table}[t]
\centering
\caption{\footnotesize{Ablation of design choices}}
\begin{tabular}{lcc}
\toprule
& Book-Caddy & Soup-Cheese \\
\midrule
\algname{} w/ 8 registers
  & \textbf{0.88 \std{0.04}} & \textbf{0.90 \std{0.02}}\\
\algname{} w/ 4 registers
  & 0.83 \std{0.05} & 0.86 \std{0.03}\\
\algname{} w/o registers
  & 0.81 \std{0.07} & 0.85 \std{0.03}\\
Cross attention \algname{}
  & 0.78 \std{0.05} & 0.86 \std{0.04} \\
\bottomrule
\end{tabular}
\label{tab:exp-ablation}
\end{table}

Results in Table \ref{tab:exp-ablation} show that adding registers to the transformer help improve the model performance. We hypothesize that adding registers facilitate the exchange of information between actions and latent image patches, which are distinct modalities. We also found that replacing AdaLN conditioning with cross attention results in worse performance. One possible explanation is that action prediction tasks benefit more from AdaLN’s global modulation than from the per-token local modulation provided by cross-attention. We note that this finding may not apply to other modalities such as language.

\subsubsection{Ablation of Learning Objectives}
To evaluate whether the performance gain of \algname{} is a result of dynamics prediction or pure reconstruction, we pretrain a \algname{} to reconstruct the \textit{current} observations instead of the \textit{future} observations. This incentivizes the model to learn about image features, but not about temporal dynamics. Table. \ref{tab:ablation-objective} shows that while reconstructing the current observations improves upon the base DP architecture with no image reconstruction, we find it advantageous to reconstruct future observations. This indicates that our model benefits from predicting dynamics rather than purely just image features. 

\begin{table}[t!]
\centering
\caption{Ablation of Learning Objectives}
\begin{tabular}{lcc}
\toprule
& Stack-Bowls & Block-Cabinet \\
\midrule
\algname{} Reconstruct Future Obs & 0.86 & 0.76 \\ 
\algname{} Reconstruct Current Obs & 0.70 & 0.66 \\
DP (No Reconstruction) & 0.48 & 0.60 \\
\bottomrule
\end{tabular}
\label{tab:ablation-objective}
\end{table}

\subsubsection{Learning from Internet videos}
\begin{figure}[t]
  \centering
  \includegraphics[width=1.0\linewidth]{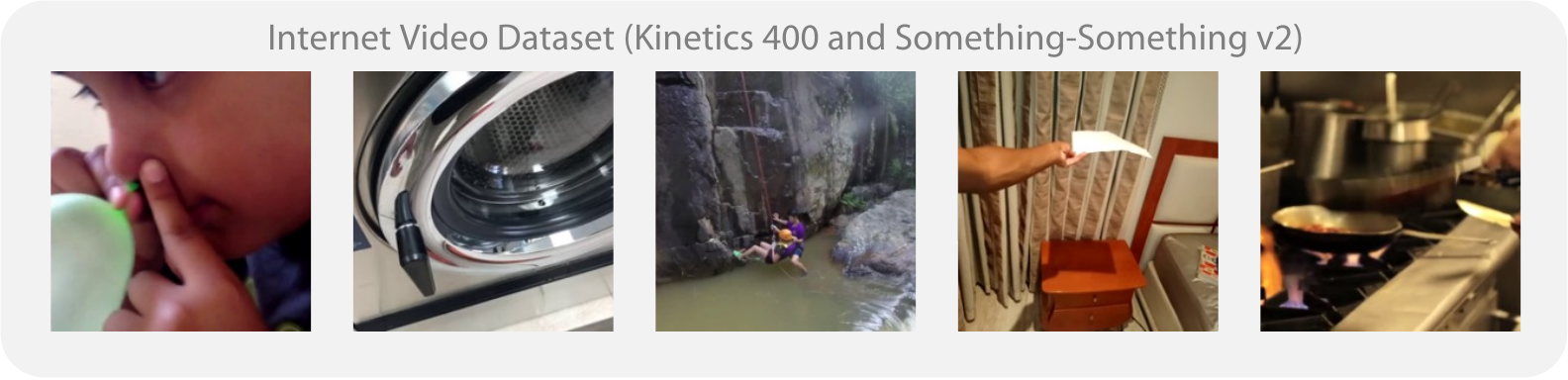}
  \caption{Visualization of Internet video dataset. We curate the dataset by combining human activity videos from Kinetics-400 \cite{carreira2017kinetics} and Something-Something-v2 \cite{goyal2017something}.}
  \label{fig:data-webvideo}
\end{figure}
\begin{table}[t!]
\centering
\caption{Cotraining on Internet Videos}
\begin{tabular}{lcc}
\toprule
& Stack-Bowls & Block-Cabinet \\
\midrule
\algname{} Robot Data + Robot Videos  & 0.92 & 0.84 \\
\algname{} Robot Data + Internet Videos & 0.88 & 0.80 \\
\algname{} Robot Data & 0.86 & 0.76 \\
\bottomrule
\end{tabular}
\label{tab:ablation-webvideo}
\end{table}
We evaluate whether \algname{} can leverage knowledge from Internet videos by including a mixture of Kinetics-400~\cite{carreira2017kinetics} and Something-Something-v2~\cite{goyal2017something} dataset in the training, which contain video clips of human activities (Fig. \ref{fig:data-webvideo}). Since the DROID setup has 3 camera views, we use random crops of the same video to impute the missing camera views. Results in Table \ref{tab:ablation-webvideo} indicate that cotraining on Internet videos shows some improvement on training only on robot data, but cotraining with in-domain robot videos still performs better. We expect these gains to be amplified in more challenging tasks and testing conditions.


\end{document}